\documentclass{article}
\usepackage{iclr2027_conference,times}
\usepackage{amsmath,amssymb}
\usepackage{graphicx,capt-of}
\graphicspath{{figures/}{figures/quantitative/}}
\usepackage{xcolor}
\usepackage{booktabs}
\usepackage{placeins}
\usepackage{hyperref}
\usepackage{url}
\hypersetup{hidelinks,
  pdftitle={What We Talk About When We Talk About Emotion Editing},
  pdfauthor={Xingzu Zhan, Lin Gu, Ruogu Fang}}

\iclrfinalcopy
\makeatletter
\renewcommand{\@maketitle}{%
  \begin{center}
    {\LARGE\scshape\@title\par}
    \vspace{14pt}
    {\normalfont\fontsize{12}{14}\selectfont\bfseries\@author\par}
    \vspace{4pt}
    \resizebox{\textwidth}{!}{%
      \begin{tabular}{@{}c@{}}
      \textsuperscript{1}Department of Biomedical Engineering, Vanderbilt University, Nashville, Tennessee, USA\\
      \textsuperscript{2}Research Institute of Electrical Communication, Tohoku University, Sendai, Japan\\
      xingzuz@andrew.cmu.edu, lin.gu@riken.jp, ruogu.fang@vanderbilt.edu
      \end{tabular}}
  \end{center}
  \vspace{6pt}}
\makeatother
\renewcommand{\headrulewidth}{0pt}
\title{What We Talk About When We Talk About Emotion Editing}
\author{Xingzu Zhan, Lin Gu\textsuperscript{2}, Ruogu Fang\textsuperscript{1}}

\begin{document}
\maketitle

\begin{center}
    \begin{minipage}{\textwidth}
        \centering
        \includegraphics[width=\linewidth]{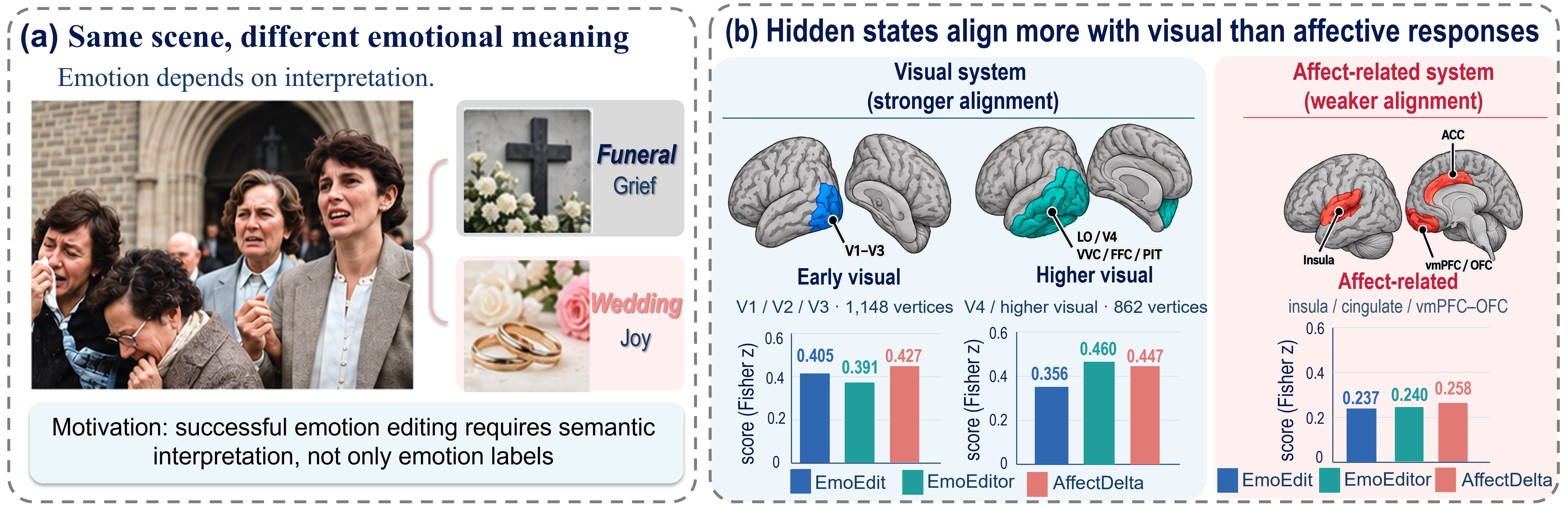}
        \captionof{figure}{Emotional image editing requires connecting visual content to emotional meaning. (a) Conceptual illustration of cognitive reappraisal: women crying outside a church may suggest grief or joy under funeral or wedding interpretations~\citep{ochsner2002rethinking}. (b) Across 400 source--target pairs, hidden states from EmoEdit, EmoEditor, and AffectDelta show stronger correspondence with TRIBE-predicted target-image responses in both early and higher visual regions than in affect-related regions (Fisher $z$ scores). Appendix~\ref{sec:baseline_hidden_correspondence} provides the encoding protocol and controls for EmoEdit and EmoEditor.}
        \label{fig:motivation}
    \end{minipage}
\end{center}

\begin{abstract}
Emotional image editing aims to change the emotion a scene conveys while preserving its essential content. Because the same scene can support different emotional interpretations, achieving an emotional goal requires visual changes grounded in the source context. Motivated by the Component Process Model, which links emotional responses to event appraisal, we investigate neural representations as a reference for these editing decisions. Using TRIBE v2, an fMRI encoding model, we analyze 400 source--target pairs across eight emotion categories. Hidden states from EmoEdit, EmoEditor, and AffectDelta correspond more strongly to predicted target-image responses in visual regions than in affect-related regions. This imbalance motivates ROI-ReFT, which combines source visual responses as a content reference with a predicted affective response change as an editing direction. Two interacting branches use both conditions to guide a frozen editor throughout generation. Predicted responses of the generated image supervise affective change and constrain visual drift. On a post hoc selected subset of 400 editing requests, ROI-ReFT achieves 91.750\% target-emotion accuracy and 83.250\% joint success, with the same output both matching the target emotion and reaching source--output DINO-I of at least 0.80. Our code and dataset will be publicly released.
\end{abstract}

\section{Introduction}
\label{sec:introduction}

Emotional image editing aims to change the affect an image conveys or evokes while preserving scene identity. EmoEdit and EmoEditor modify semantic content and contextual cues~\citep{yang2025emoedit,lin2025make}. AffectDelta specifies source-to-target changes in emotion distributions~\citep{zhan2026affectdelta}, while EmoScope explores image-specific editing strategies~\citep{li2026emoscope}. An emotional target specifies the intended outcome, while realizing it requires choosing visual changes appropriate to the source scene.

This choice depends on how visual content is interpreted. In their cognitive reappraisal study, \citet{ochsner2002rethinking} illustrate reinterpretation with women crying outside a church: the scene could represent a wedding rather than a funeral. The same tears can thus suggest joy or grief, depending on the event the viewer believes is taking place (Figure~\ref{fig:motivation}a). Across the task, reinterpreting unchanged photographs reduced negative affect. This ambiguity raises the question of whether image editors capture the emotional meaning of the source and target images alongside their visual content. Emotional meaning involves interacting cognitive and bodily processes. The Component Process Model (CPM) explains how appraisals of an event's relevance, implications, coping potential, and normative significance coordinate motivation, physiology, expression, and subjective feeling~\citep{scherer2009emotions}. This account highlights the complexity of relating visual content to an emotional response and motivates examining the affective information represented by an editor.

We therefore seek a neural reference for how the source scene may be emotionally interpreted by viewers. \citet{leitao2020computational} combined fMRI with behavioral and physiological measures to identify brain networks associated with four CPM components and their dynamic synchronization. Visual-emotion studies also show distributed emotion information across visual and transmodal regions~\citep{kragel2019emotion,horikawa2020neural}. We use TRIBE v2, an fMRI encoding model, to predict cortical responses to the source image~\citep{dascoli2026foundation}. These predicted responses provide a reference for its affective context and visual content. Desired changes in affect-related responses can then guide the emotional direction of editing, while source visual responses guide content preservation. This joint guidance aims to produce an image that evokes the target emotion while retaining the source scene's principal content. Across 400 source--target pairs balanced across eight source and target emotion categories, hidden states from EmoEdit, EmoEditor, and AffectDelta show stronger correspondence with TRIBE-predicted target-image responses in both early and higher visual regions than in affect-related regions (Figure~\ref{fig:motivation}b). This pattern also holds for AffectDelta, our base editor, motivating explicit guidance for the desired affective change. Since emotional change is realized through visual cues, we combine this affective direction with a source visual reference to coordinate emotional change and content preservation.

Building on this observation and prior work on neural guidance~\citep{li2019learning,xiao2026beyond}, we introduce ROI-ReFT, a dual-ROI representation intervention for emotional image editing. It couples a source visual reference with a predicted affective response change, allowing the source scene and emotional goal to jointly guide generation. Our contributions are threefold:
\begin{itemize}
    \item We characterize visual and affective correspondence in three emotion editors, identifying stronger alignment of their hidden states with TRIBE-predicted target-image responses in visual regions than in affect-related regions.
    \item We propose a dual-ROI intervention with two interacting branches that read both neural conditions throughout generation. Supervision through the generated image's predicted responses trains their collaboration to realize affective change while constraining visual drift.
    \item On a post hoc selected subset of 400 editing requests, ROI-ReFT achieves 91.750\% target-emotion accuracy and 83.250\% joint success at DINO-I $\geq 0.80$, improving emotional accuracy and all three preservation metrics over AffectDelta. Ablations further support the benefits of interaction and dual neural supervision.
\end{itemize}

\section{Related Work}
\label{sec:related_work}

\subsection{Emotional Image Editing}
\label{sec:related_emotional_editing}

Emotional image editing studies how visual changes alter the emotions an image evokes. Early work linked affective words to color themes for image recoloring~\citep{wang2013affective} and transferred color and texture toward emotion distributions associated with a reference image~\citep{peng2015mixed}. These approaches established appearance attributes and emotion distributions as useful controls for affective transformation. Their emphasis on color and texture offers direct control over visual atmosphere; later semantic editing methods expand the available changes to the depicted objects and situations.

Diffusion-based methods extend these controls to semantic content and scene context. EmoEdit combines an emotion adapter with an instruction loss to learn context-dependent semantic modifications~\citep{yang2025emoedit}. EmoEditor integrates global emotional and local visual cues, conditions editing on the difference between a target emotion code and the predicted source distribution, and refines generation through iterative emotion inference~\citep{lin2025make}. AffectDelta represents the requested change as a signed difference between full source and target emotion distributions, including transitions that retain the dominant category~\citep{zhan2026affectdelta}. EmoScope explores image-specific editing strategies through affordance reasoning and iterative verification of emotional plausibility and consistency~\citep{li2026emoscope}. Together, these methods develop how emotional goals are specified and translated into scene changes. Our approach adds predicted neural responses as editing guidance: source visual responses provide a content reference, while desired changes in affect-related responses provide an editing direction.

\subsection{Theoretical Models of Emotion}
\label{sec:related_emotion_theory}

Emotion theories differ in how they characterize emotional states and explain their generation. Basic-emotion accounts distinguish families of emotions with characteristic response patterns~\citep{ekman1992argument}, whereas dimensional models organize affect along continuous axes such as pleasure and arousal~\citep{russell1980circumplex}. The theory of constructed emotion emphasizes how past experience, interoceptive prediction, and conceptual knowledge shape emotional experiences~\citep{barrett2017theory}.

Appraisal theories relate emotional differences to evaluations of an event's significance, including certainty, responsibility, and control~\citep{smith1985patterns}. The Component Process Model (CPM) develops this perspective into a dynamic account: appraisals drive coordinated changes in motivation, physiology, expression, and subjective feeling as an event unfolds~\citep{scherer2009emotions}. This makes the relation between an event and the viewer central to explaining emotional change. Computational models such as EMA connect appraisal and coping to an evolving interpretation of the situation~\citep{marsella2009ema}. Emotion-regulation models further distinguish interventions at different stages of emotion generation~\citep{gross1998emerging}, with cognitive reappraisal studies demonstrating that changing an interpretation can alter affect even when the image remains unchanged~\citep{ochsner2002rethinking}.

We adopt CPM as the main theoretical motivation for relating scene interpretation to emotional change. Neuroimaging work has identified networks associated with CPM components and their dynamic coordination~\citep{leitao2020computational}, while studies of visually evoked emotion report representations across visual and transmodal regions~\citep{kragel2019emotion,horikawa2020neural}. Neural information has also served as a learning signal: \citet{li2019learning} regularize visual models using neural representational similarities, while \citet{xiao2026beyond} use fMRI-guided interventions and fine-tuning to improve language-model reasoning. These studies provide a methodological precedent for using neural representations to guide model behavior~\citep{chen2026calibrated}, complementing their role in assessing correspondence. These findings motivate drawing on distributed neural information for editing. Our formulation uses predicted visual and affect-related responses jointly to guide content preservation and emotional change.

\section{Method}
\label{sec:method}

\subsection{Overview}
\label{sec:method_overview}

Our analysis in the Introduction shows that existing editors' hidden states correspond more strongly to TRIBE-predicted visual responses than to affect-related responses (Figure~\ref{fig:motivation}b). This imbalance motivates explicitly guiding the generation process with the desired affective response change. Because emotional changes are expressed through scene-dependent visual cues, we pair this direction with a source visual reference. Given a source image $I_s$ and supplied source and target emotion distributions $p_s,p_t\in\mathbb{R}^{8}$, we use these two neural conditions to generate an edited image $I_g$ that conveys the requested emotion while retaining the scene's principal content.

\begin{figure}[tb]
\centering
\includegraphics[width=\linewidth]{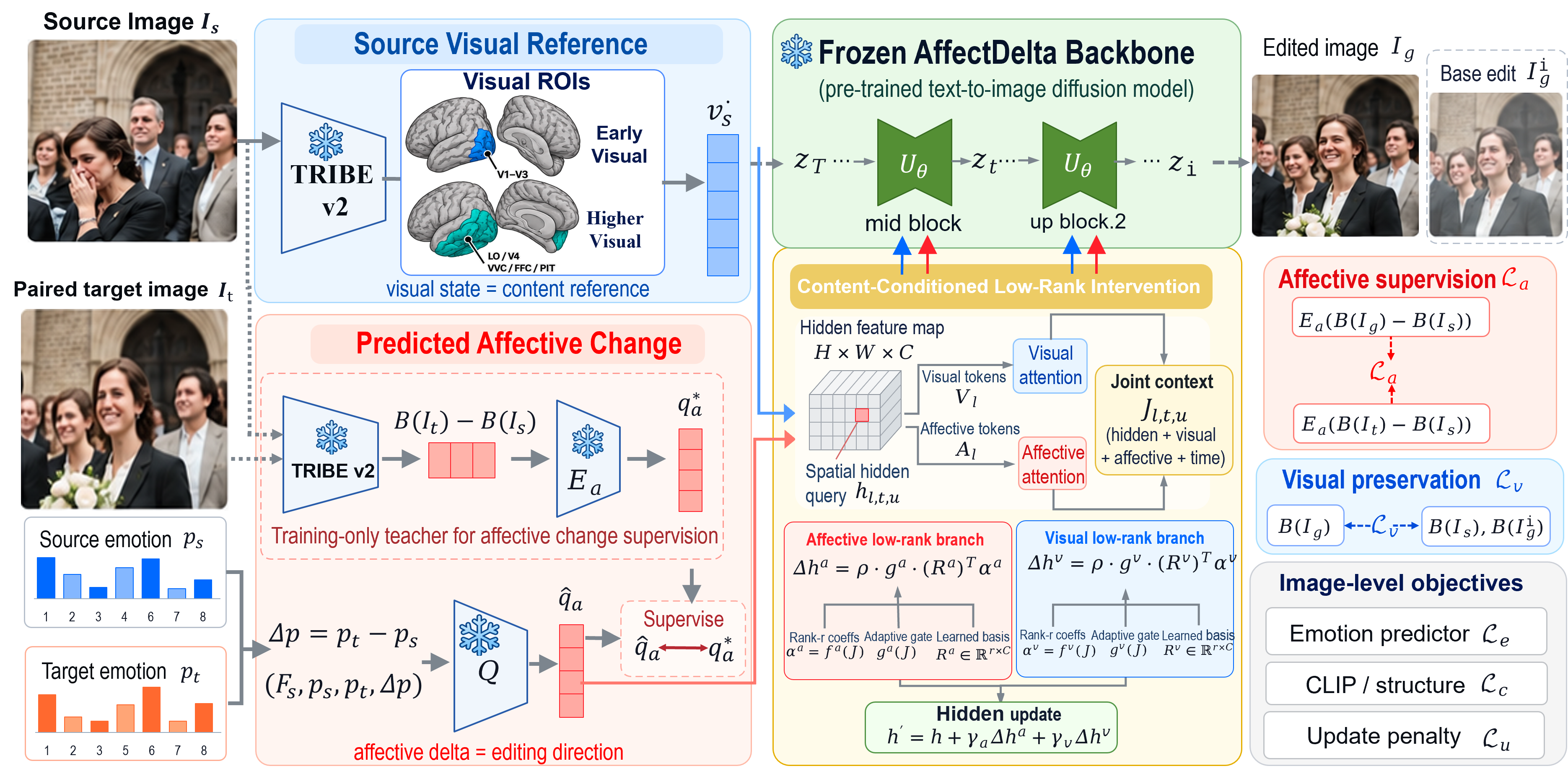}
\caption{Overview of joint visual and affective neural guidance. TRIBE-predicted source visual responses provide a content reference, while a source-conditioned predictor supplies an affective editing direction. Both conditions jointly drive gated low-rank updates throughout the frozen diffusion editor. Predicted responses of the generated image supervise affective change and constrain visual drift, complemented by auxiliary objectives. Paired target responses supervise predictor pretraining and generated affective change during training. The predictor is frozen during intervention training; inference uses the source image and emotion conditions. Snowflakes denote frozen modules during intervention training.}
\label{fig:pipeline}
\end{figure}

Figure~\ref{fig:pipeline} shows how we turn these two roles into editing guidance. We build on the frozen AffectDelta editor~\citep{zhan2026affectdelta}, retaining its source-image latent and emotion conditioning, and use a frozen TRIBE v2 encoding model $B$ to predict population cortical responses~\citep{dascoli2026foundation}. We first construct a source visual response state and predict a source-relative affective response change. These conditions jointly regulate the editor's hidden states throughout denoising. We then train their combined action through the generated image: its predicted responses supervise affective change and constrain visual drift. A paired target image $I_t$ supplies training supervision; inference uses the source image and emotion conditions.

\subsection{Constructing Neural Guidance}
\label{sec:neural_conditions}

The two conditions describe different aspects of an edit: the visual condition specifies the scene that should remain recognizable, while the affective condition specifies the emotional transition to realize. We therefore encode the former as a source response state and the latter as a signed response change, corresponding to the two guidance paths on the left of Figure~\ref{fig:pipeline}.

For the visual reference, we encode the source image's predicted responses in visual regions of interest (ROIs):
\begin{equation}
v_s=E_v^{\mathrm{state}}(B(I_s))\in\mathbb{R}^{32}.
\label{eq:visual_state}
\end{equation}
The frozen state encoder $E_v^{\mathrm{state}}$ summarizes two visual ROI groups into 16 coordinates each. This source-dependent state provides a content reference throughout generation. The original source latent remains a separate spatial input to the editor, preserving layout information alongside the neural reference.

For affective guidance, the required change depends on both the source scene and the requested emotional transition. A training pair provides the target change $q_a^*=E_a(B(I_t)-B(I_s))$. The frozen encoder $E_a$ summarizes three affect-related ROI groups into eight coordinates each, preserving zero change and the sign of the response difference. This representation specifies how the response should move relative to the source. To supply this direction without a paired target image at inference, we predict it from source-image features $F_s$ and the emotion conditions:
\begin{equation}
\widehat q_a=Q(F_s,p_s,p_t,p_t-p_s)\in\mathbb{R}^{24}.
\label{eq:affective_prediction}
\end{equation}
Here $F_s$ is pooled from the source image's VAE latent. We pretrain $Q$ on paired affective change codes and freeze it during intervention training. Its prediction $\widehat q_a$ is the affective condition in both training and inference, while paired target responses supervise the generated change during training. Thus, the predictor-supervision path in Figure~\ref{fig:pipeline} belongs to pretraining, and the output-supervision path trains the intervention.

Both $v_s$ and $\widehat q_a$ are computed once per edit and reused throughout denoising. Appendix~\ref{sec:neural_encoding_details} details the group encodings, normalization conventions, source features, and predictor pretraining.

\subsection{Joint Neural Intervention}
\label{sec:dual_intervention}

The two guidance signals must act together because the visual changes that express an emotion depend on the source scene. Conversely, preserving the scene must allow the changes needed to convey that emotion. We therefore let both intervention branches read both conditions: the visual reference informs affective editing, and the affective direction informs visual preservation. Following representation-level adaptation in ReFT~\citep{wu2024reft}, we implement this interaction through gated low-rank updates to the frozen diffusion editor, as shown in the central module of Figure~\ref{fig:pipeline}.

At layer $\ell$, timestep $t$, and spatial position $u$, the current hidden vector $h_{\ell,t,u}$ provides a spatial query for separately reading the visual and affective ROI tokens. The retrieved information is combined with source features, time and layer context, and a direct projection of $[p_s,p_t,p_t-p_s]$. This joint context gives both branches access to the current scene representation, both neural conditions, and the requested emotional transition. The direct emotion projection is trainable and complements the frozen predictor $Q$.

Each branch uses its own action network, low-rank basis, and sigmoid gate to translate the joint context into a hidden update. The two independently gated updates are added to the current representation:
\begin{equation}
h'_{\ell,t,u}=h_{\ell,t,u}
+\gamma_a\delta h^a_{\ell,t,u}
+\gamma_v\delta h^v_{\ell,t,u},
\label{eq:dual_write}
\end{equation}
where $\delta h^a$ and $\delta h^v$ denote the affective and visual updates, and $\gamma_a,\gamma_v$ scale their strengths. The gates can activate simultaneously, allowing both branches to contribute at the same position and step. Spatial variation follows from the hidden-state queries, while the shared context coordinates the two branches' actions.

We apply both branches at two U-Net scales, \texttt{mid\_block} and \texttt{up\_blocks.2}, at every step of the 20-step DDIM trajectory~\citep{song2021denoising}. Although $v_s$ and $\widehat q_a$ remain fixed, the evolving hidden states and noise level change the queries, actions, and gates across the trajectory. The intervention thus responds to the current generation state at both scales. The backbone remains frozen while the intervention learns how to combine the affective direction with the source visual reference. Appendix~\ref{sec:intervention_details} specifies the attention, context construction, gating, low-rank parameterization, initialization, and layer configuration.

\subsection{Supervision through the Generated Image}
\label{sec:output_supervision}

The two conditions guide how the editor changes its representations; training must connect those changes to the image they produce. We therefore supervise the completed output along the same two dimensions: the affective change it realizes and the visual content it retains. After the full generation trajectory, the frozen, differentiable TRIBE model predicts the output image's cortical responses. Losses on these responses propagate through the image decoder and all denoising steps to the intervention parameters, implementing the output-supervision paths on the right of Figure~\ref{fig:pipeline}.

For affective supervision, we align the generated response change with the change specified by the paired target:
\begin{equation}
\mathcal{L}_a=\frac{1}{24}
\left\|E_a(B(I_g)-B(I_s))-E_a(B(I_t)-B(I_s))\right\|_2^2.
\label{eq:affective_loss}
\end{equation}
Source, target, and generated responses are computed with the same continuous-image teacher. The paired target anchors the desired output change, while $\widehat q_a$ supplies the editor's input guidance. This keeps output supervision independent of the predictor's estimate.

Visual supervision limits source-relative neural drift while allowing the visual modifications needed for the emotional goal. We use a soft budget relative to a matched edit from the frozen base model. Let $I_g^0$ be the output with the same source, emotion conditions, initial noise, and sampler, with interventions disabled. If $d_v(I,I_s)$ measures source-relative response differences in the visual ROI codes, then
\begin{equation}
\mathcal{L}_v=
\left[d_v(I_g,I_s)-\operatorname{sg}(d_v(I_g^0,I_s))-\kappa_v\right]_+^2,
\label{eq:visual_budget}
\end{equation}
where $[x]_+=\max(x,0)$ and $\operatorname{sg}$ stops gradients. The matched base edit provides a reference amount of visual change under the same editing request. The loss penalizes only deviation beyond this reference plus allowance $\kappa_v$, permitting necessary scene changes while limiting additional drift introduced by the intervention.

We complement these neural objectives with an emotion loss $\mathcal{L}_e$ for the requested distribution and a content loss $\mathcal{L}_c$ for image-level preservation. The content loss separately budgets CLIP deviation from the matched base output and structural deviation from the source image. We also retain the editor's noise-prediction objective $\mathcal{L}_{\epsilon}$ and regularize the combined hidden update with $\mathcal{L}_u$. The full objective is
\begin{equation}
\mathcal{L}=\mathcal{L}_{\epsilon}
+\lambda_a\mathcal{L}_a+\lambda_v\mathcal{L}_v
+\lambda_e\mathcal{L}_e+\lambda_c\mathcal{L}_c
+\lambda_u\mathcal{L}_u.
\label{eq:training_objective}
\end{equation}
Together, these objectives train the joint intervention to translate an affective direction into changes grounded in the source scene. Appendix~\ref{sec:loss_details} specifies the distance functions and auxiliary losses, and Appendix~\ref{sec:training_inference} provides the training and inference settings.

\section{Experiments}
\label{sec:experiments}
We evaluate whether joint neural guidance improves emotional editing and source-content preservation, identify the contributions of its components, and examine whether it strengthens affective correspondence in hidden representations.

\subsection{Experimental Setup}
\paragraph{Data.}
We evaluate all methods on the same post hoc selected subset of 400 editing requests from AffectPair-249K, introduced by AffectDelta~\citep{zhan2026affectdelta}. Source and target categories are each balanced across amusement, anger, awe, contentment, disgust, excitement, fear, and sadness, with 50 examples per category in each distribution.

\paragraph{Baselines.}
We compare seven methods spanning general editing and emotion-oriented transformation. LDAST~\citep{fu2022language} transfers artistic style from language, SDEdit~\citep{meng2022sdedit} edits through noise addition and diffusion denoising, and InstructPix2Pix (IP2P)~\citep{brooks2023instructpix2pix} follows natural-language editing instructions. Among emotion-oriented methods, AIF~\citep{weng2023affective} maps textual emotions to color and texture changes; EmoEdit~\citep{yang2025emoedit} learns emotion-conditioned semantic modifications; and EmoEditor~\citep{lin2025make} combines global emotional context with local visual cues. AffectDelta~\citep{zhan2026affectdelta} conditions editing on signed differences between source and target emotion distributions. As our base editor, it also provides a direct comparison for assessing the gains from neural intervention.

\paragraph{Metrics.}
A shared, frozen eight-class ResNet-18 emotion predictor from EmoEditor~\citep{lin2025make} measures target-emotion accuracy (Emo-A) and the signed increase in target-emotion probability ($\Delta$Target). DINO-I, CLIP-I, and LPIPS assess preservation relative to the source. Joint Success Rate (JSR) measures the fraction of all outputs that both match the target emotion and reach a DINO-I threshold $\tau$, testing whether the same edit satisfies both goals. Appendix~\ref{sec:evaluation_metrics} gives the metric definitions and evaluator configurations; Appendix~\ref{sec:training_inference} provides training and inference details.

\subsection{Comparison with State-of-the-Art Methods}
\paragraph{Quantitative Comparison.}
ROI-ReFT improves emotional control and content preservation together (Table~\ref{tab:quantitative-comparison}). It achieves the highest Emo-A of 91.750\%, exceeding AffectDelta's 70.500\% by 21.250 percentage points, and the highest $\Delta$Target of $+0.301$ versus $+0.199$. Relative to AffectDelta, DINO-I rises from 0.749 to 0.862, CLIP-I from 0.751 to 0.819, and LPIPS decreases from 0.390 to 0.252. ROI-ReFT ranks first on CLIP-I and LPIPS and second on DINO-I. IP2P obtains higher DINO-I (0.948), but its Emo-A is 53.500\%, illustrating why visual similarity must be considered alongside emotional correctness. The simultaneous gains over the base editor support combining an affective editing direction with a source visual reference.

\begin{table*}[t]
\centering
\caption{Quantitative comparison of emotion editing and image preservation on the selected 400 editing requests. Arrows indicate whether higher or lower values are better. Best and second-best results in each column are bold and underlined, respectively, with ranks determined before rounding. Ties receive the same marking.}
\label{tab:quantitative-comparison}
\begingroup
\small
\setlength{\tabcolsep}{4pt}
\renewcommand{\arraystretch}{1.1}
\begin{tabular*}{0.95\linewidth}{@{\extracolsep{\fill}}lccccc@{}}
\toprule
Method & Emo-A $\uparrow$ & $\Delta$Target $\uparrow$ & DINO-I $\uparrow$ & CLIP-I $\uparrow$ & LPIPS $\downarrow$ \\
\midrule
LDAST~\citep{fu2022language} & 30.500\% & $-0.154$ & 0.508 & 0.672 & 0.462 \\
SDEdit~\citep{meng2022sdedit} & 22.250\% & $-0.229$ & 0.159 & 0.544 & 0.565 \\
IP2P~\citep{brooks2023instructpix2pix} & 53.500\% & $+0.039$ & \textbf{0.948} & 0.739 & 0.270 \\
AIF~\citep{weng2023affective} & 31.250\% & $-0.141$ & 0.774 & \underline{0.806} & \underline{0.267} \\
EmoEdit~\citep{yang2025emoedit} & 66.750\% & $+0.169$ & 0.745 & \underline{0.806} & 0.267 \\
EmoEditor~\citep{lin2025make} & 45.750\% & $-0.032$ & 0.807 & 0.795 & 0.267 \\
AffectDelta~\citep{zhan2026affectdelta} & \underline{70.500\%} & $\underline{+0.199}$ & 0.749 & 0.751 & 0.390 \\
\midrule
\textbf{ROI-ReFT (Ours)} & \textbf{91.750\%} & $\mathbf{+0.301}$ & \underline{0.862} & \textbf{0.819} & \textbf{0.252} \\
\bottomrule
\end{tabular*}
\vspace{-4mm}
\endgroup
\end{table*}

\begin{figure}[!htbp]
\centering
\includegraphics[width=0.90\linewidth]{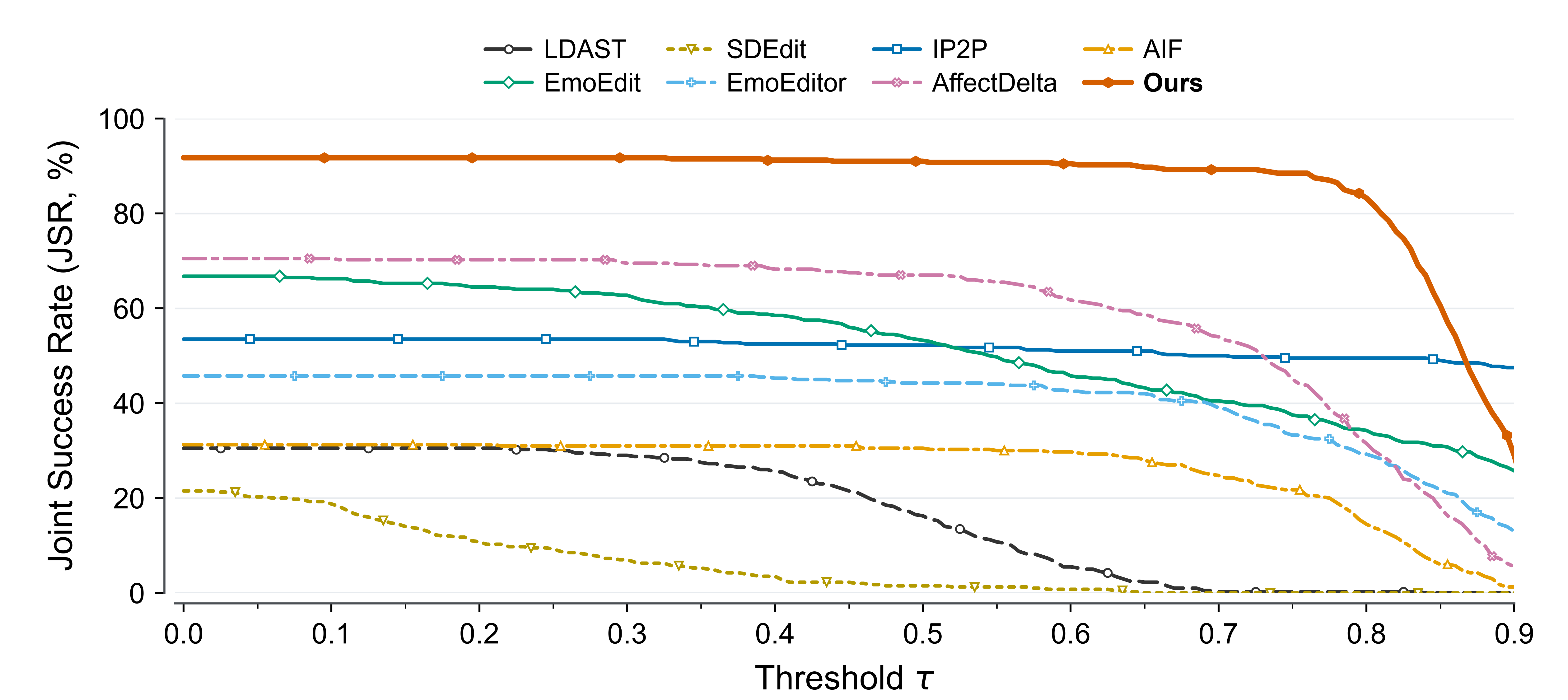}
\vspace{-5mm}
\caption{Joint Success Rate (JSR) across DINO-I thresholds on the same 400 selected editing requests. An output counts as successful only when its predicted emotion matches the target and its source--output DINO-I is at least $\tau$. Higher curves indicate more outputs satisfying both requirements. ROI-ReFT achieves 83.250\% JSR at $\tau=0.80$ (333 of 400 outputs).}
\label{fig:jsr-comparison}
\vspace{-5mm}
\end{figure}

\paragraph{Joint Emotional and Visual Success.}
JSR tests whether the aggregate gains above occur in the same outputs. Figure~\ref{fig:jsr-comparison} shows that ROI-ReFT leads across most preservation thresholds, reaching 83.250\% at $\tau=0.80$ (333 of 400 outputs). IP2P preserves source content closely and leads at the strictest thresholds, but its lower emotional accuracy limits the fraction of edits that achieve the requested emotion. ROI-ReFT's higher joint success across most thresholds supports coordinating emotional change with visual preservation.

\begin{figure*}[t]
\centering
\begingroup
\edef\qualcellwidth{\the\dimexpr(0.99\linewidth-1.35em-0.5pt-12pt)/9\relax}
\setlength{\tabcolsep}{0pt}
\renewcommand{\arraystretch}{0.8}
\definecolor{qualTargetWarm}{RGB}{175,70,15}
\definecolor{qualTargetCool}{RGB}{0,97,153}
\newcommand{\qualimage}[1]{%
  \parbox[c][\qualcellwidth][c]{\qualcellwidth}{%
    \centering
    \includegraphics[width=\linewidth,height=\linewidth,keepaspectratio]{#1}%
  }%
}
\newcommand{\qualtarget}[2]{%
  \parbox[c][\qualcellwidth][c]{1.35em}{%
    \centering
    \rotatebox[origin=c]{90}{%
      \textcolor{#1}{\scriptsize\rmfamily\bfseries\itshape ``#2''}%
    }%
  }%
}
\begin{tabular}{@{}c@{\hspace{0.5pt}}c*{8}{@{\hspace{1.5pt}}c}@{}}
{} &
{\scriptsize\bfseries Input} &
{\scriptsize\bfseries Ours} &
{\scriptsize\bfseries AffectDelta} &
{\scriptsize\bfseries EmoEditor} &
{\scriptsize\bfseries EmoEdit} &
{\scriptsize\bfseries AIF} &
{\scriptsize\bfseries IP2P} &
{\scriptsize\bfseries SDEdit} &
{\scriptsize\bfseries LDAST}
\\[1pt]

\qualtarget{qualTargetWarm}{Amusement} &
\qualimage{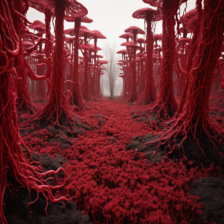} &
\qualimage{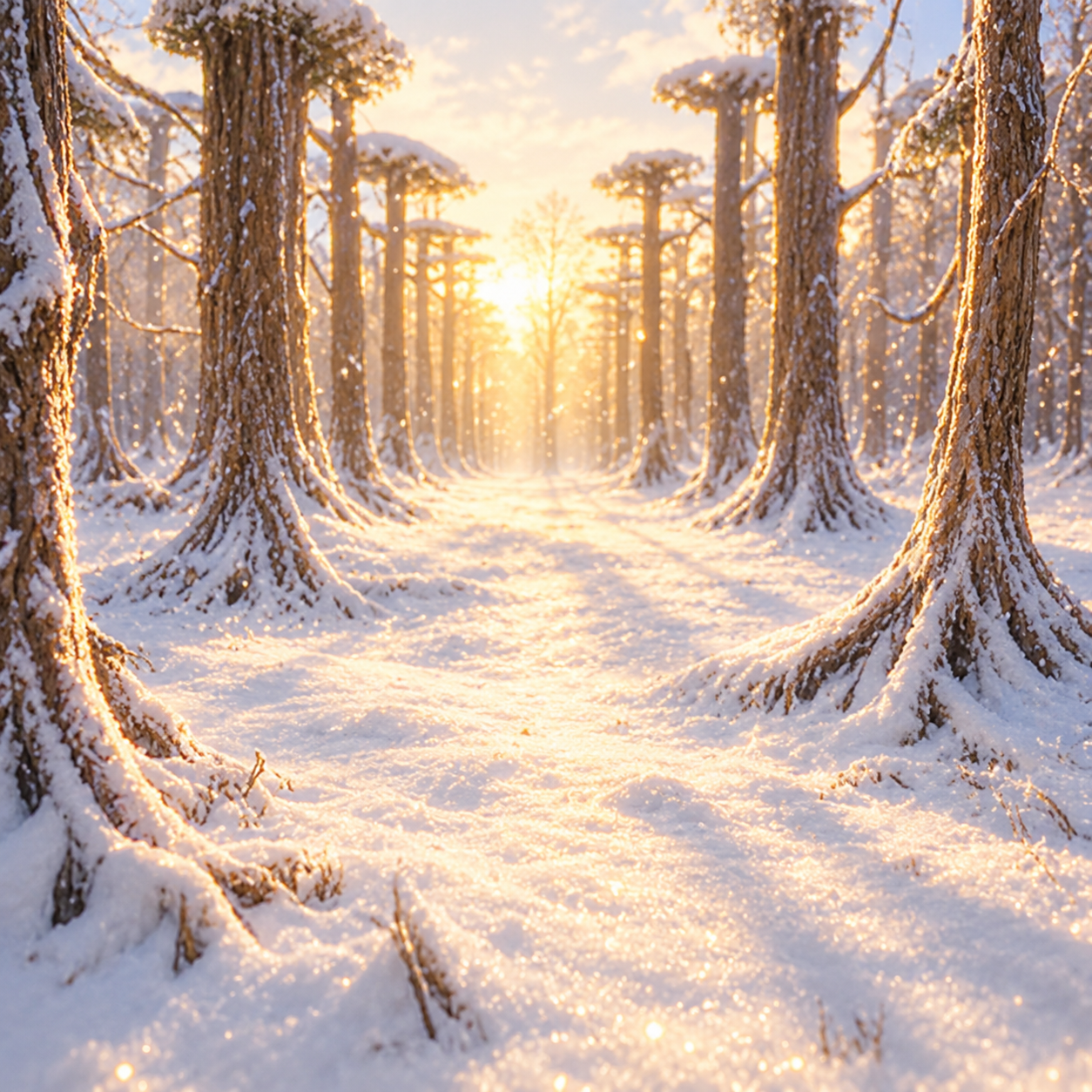} &
\qualimage{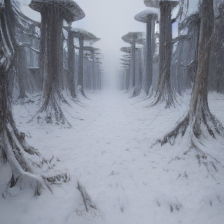} &
\qualimage{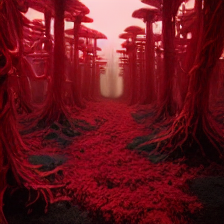} &
\qualimage{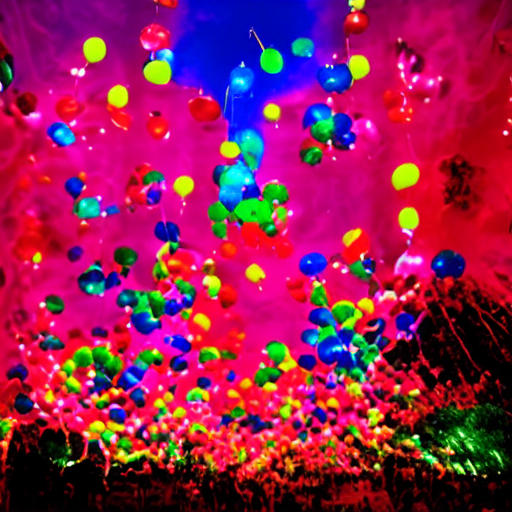} &
\qualimage{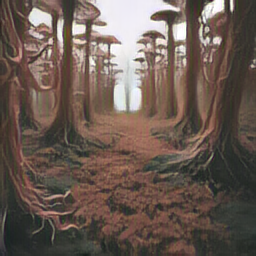} &
\qualimage{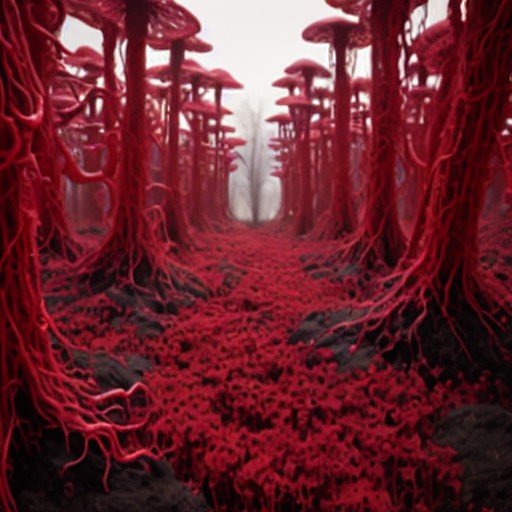} &
\qualimage{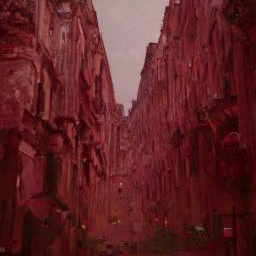} &
\qualimage{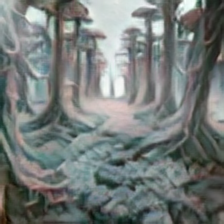}
\\
\noalign{\vskip1.5pt}

\qualtarget{qualTargetCool}{Anger} &
\qualimage{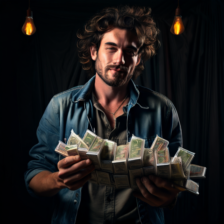} &
\qualimage{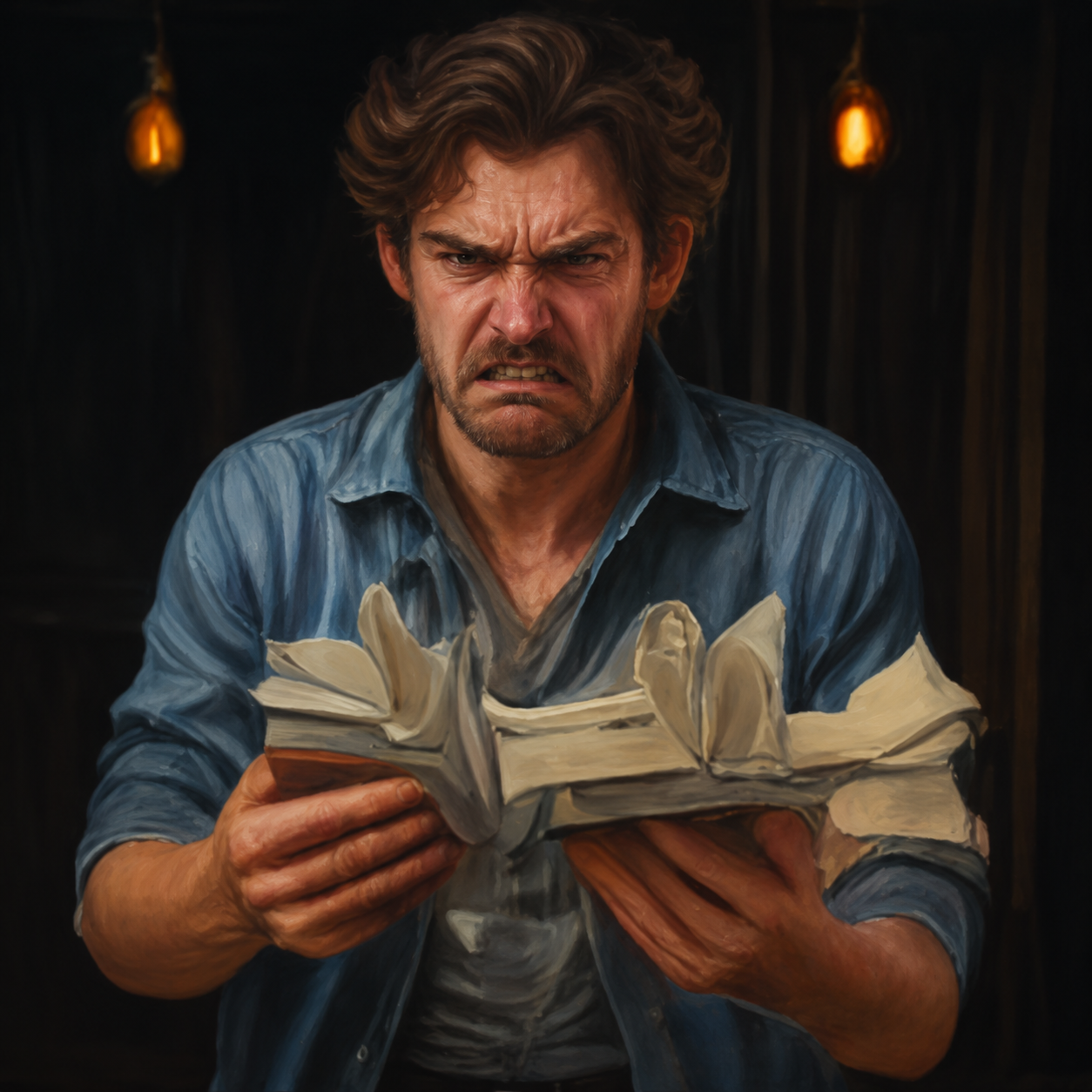} &
\qualimage{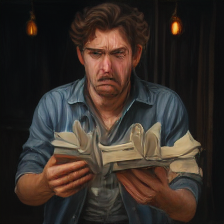} &
\qualimage{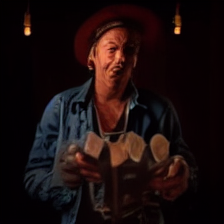} &
\qualimage{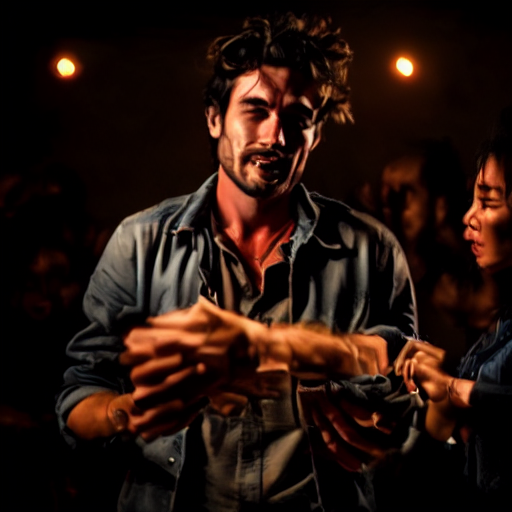} &
\qualimage{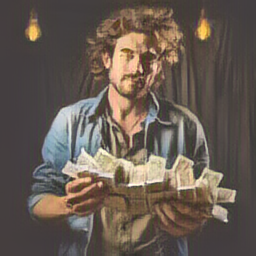} &
\qualimage{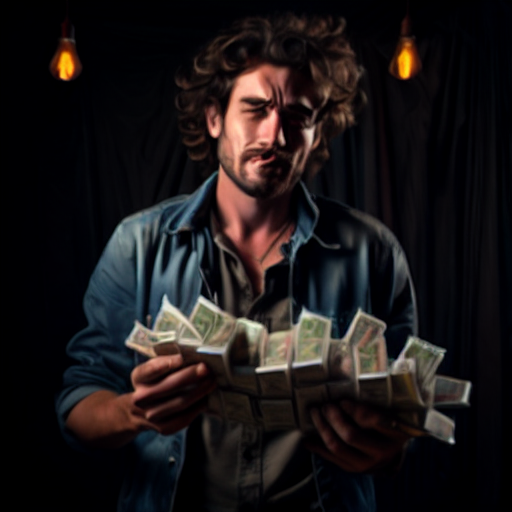} &
\qualimage{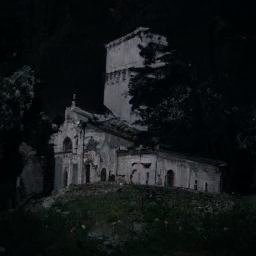} &
\qualimage{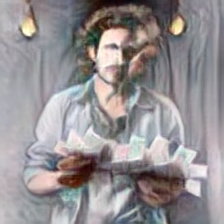}
\\
\noalign{\vskip1.5pt}

\qualtarget{qualTargetWarm}{Awe} &
\qualimage{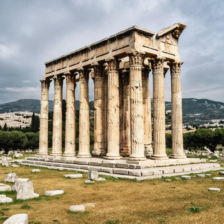} &
\qualimage{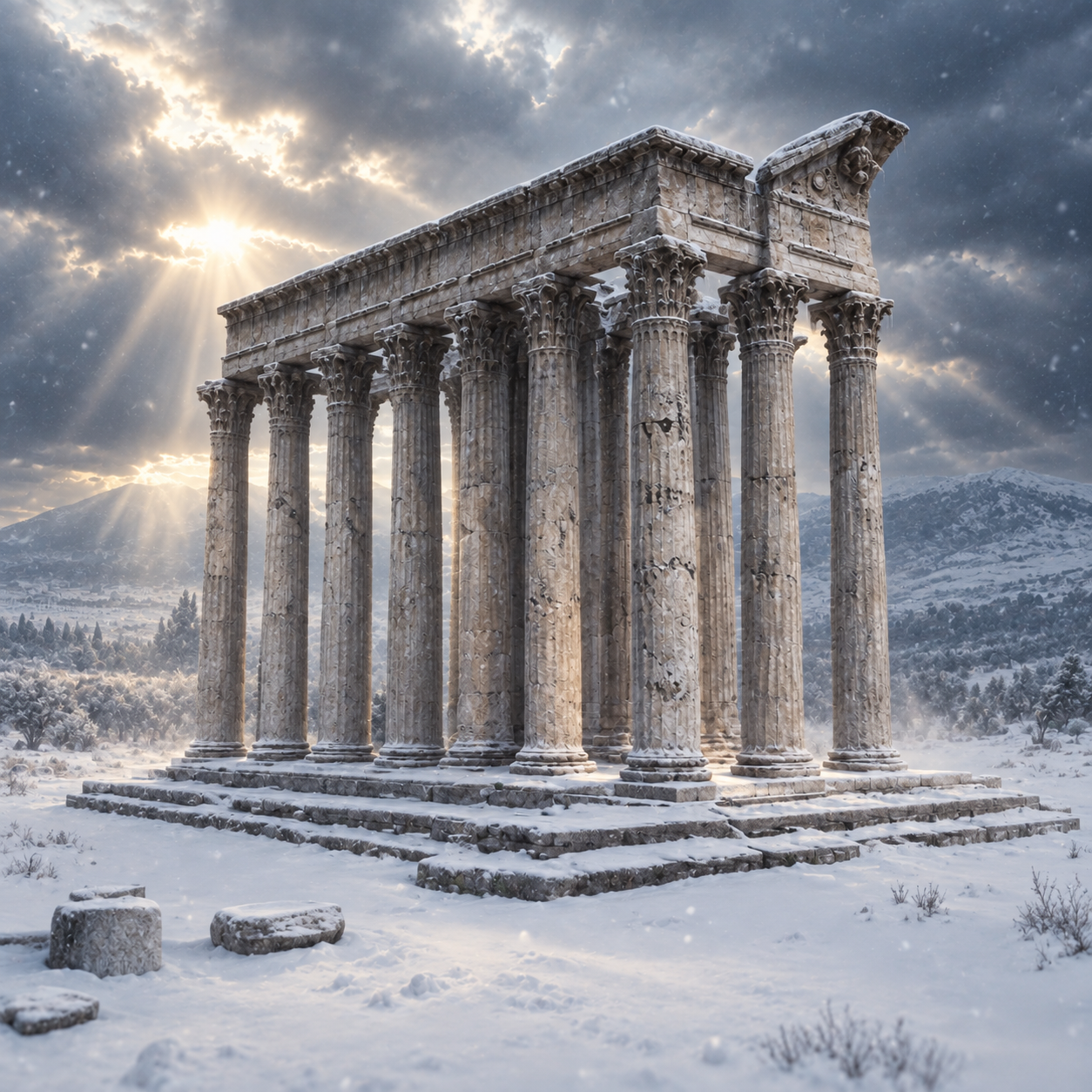} &
\qualimage{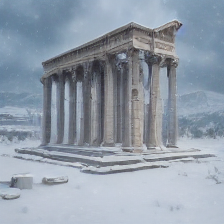} &
\qualimage{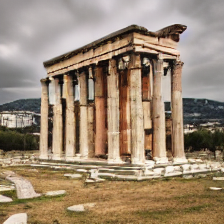} &
\qualimage{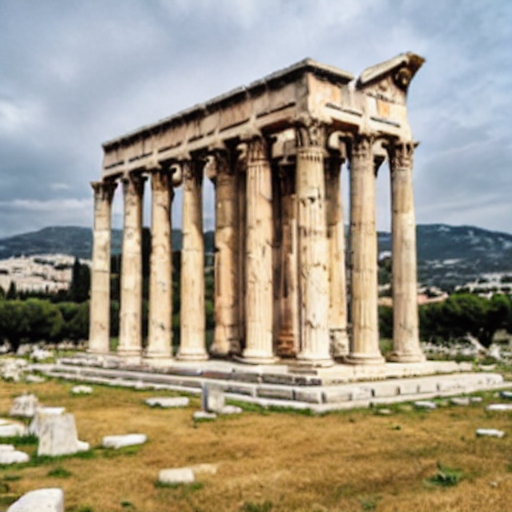} &
\qualimage{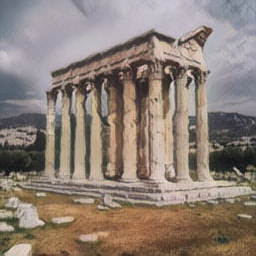} &
\qualimage{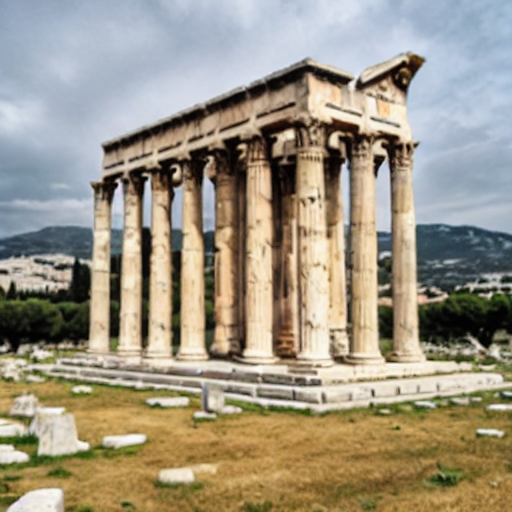} &
\qualimage{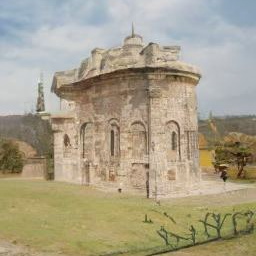} &
\qualimage{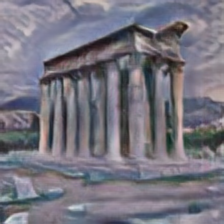}
\\
\noalign{\vskip1.5pt}

\qualtarget{qualTargetCool}{Sadness} &
\qualimage{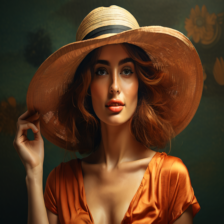} &
\qualimage{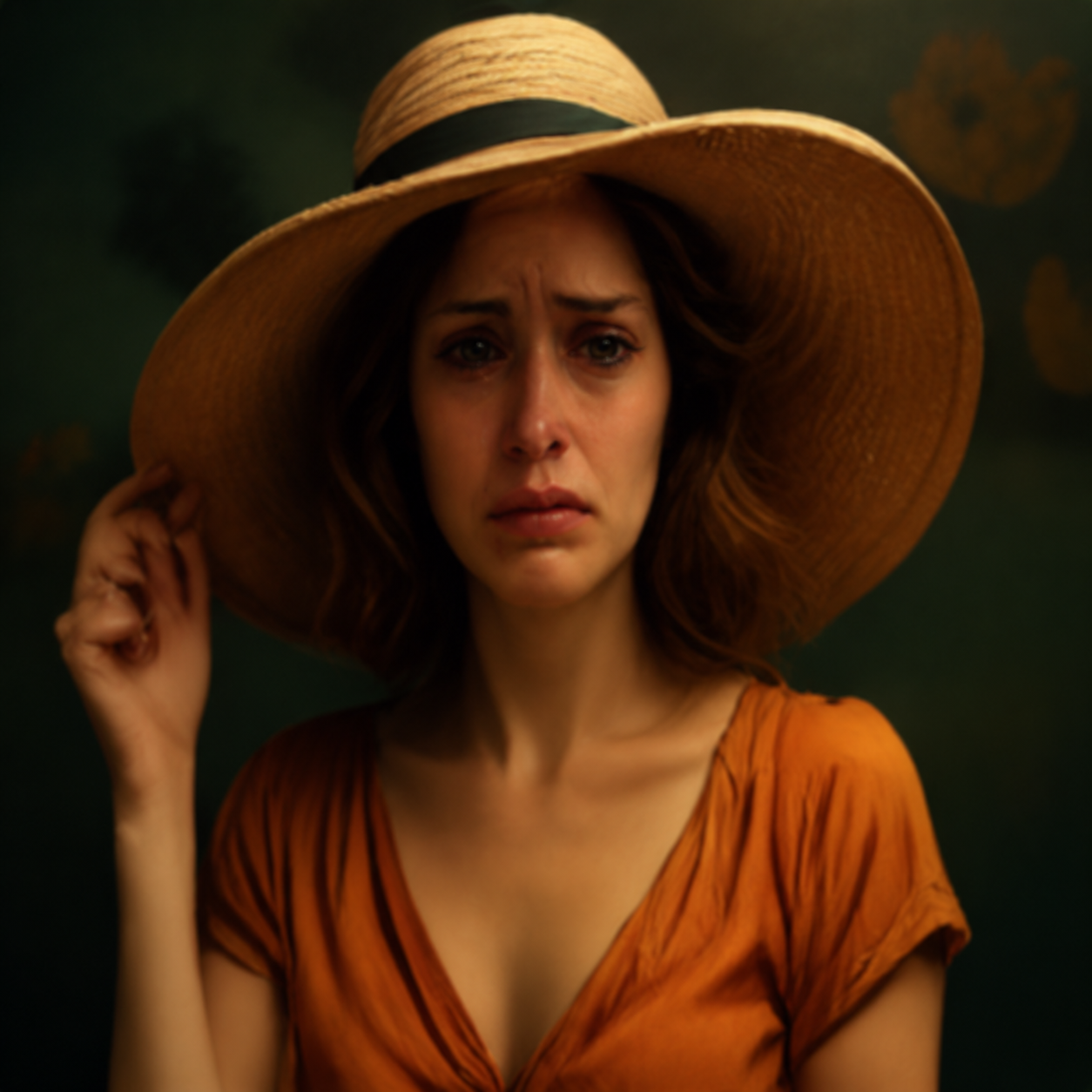} &
\qualimage{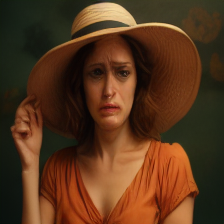} &
\qualimage{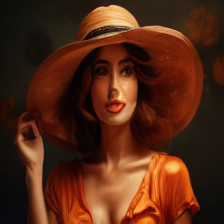} &
\qualimage{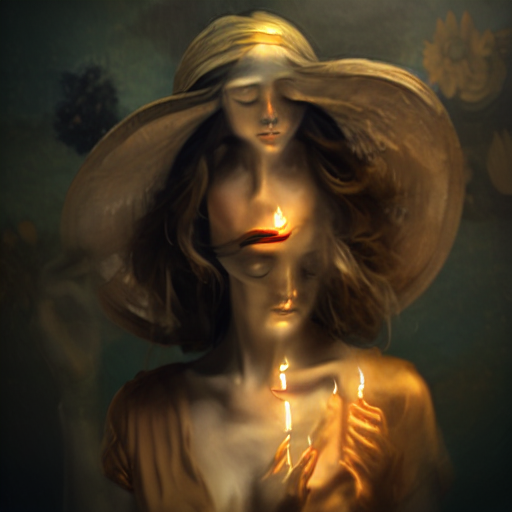} &
\qualimage{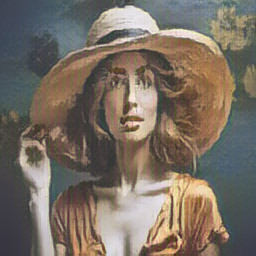} &
\qualimage{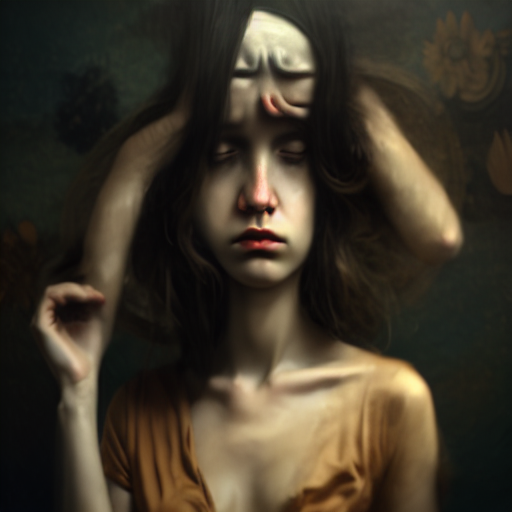} &
\qualimage{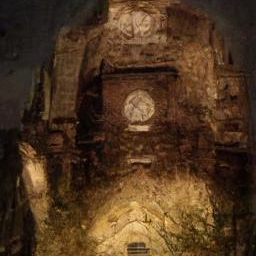} &
\qualimage{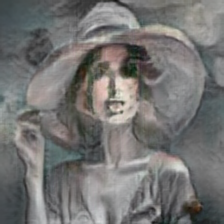}
\\
\noalign{\vskip1.5pt}

\qualtarget{qualTargetWarm}{Excitement} &
\qualimage{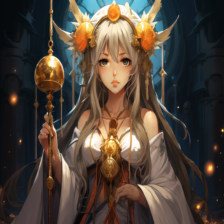} &
\qualimage{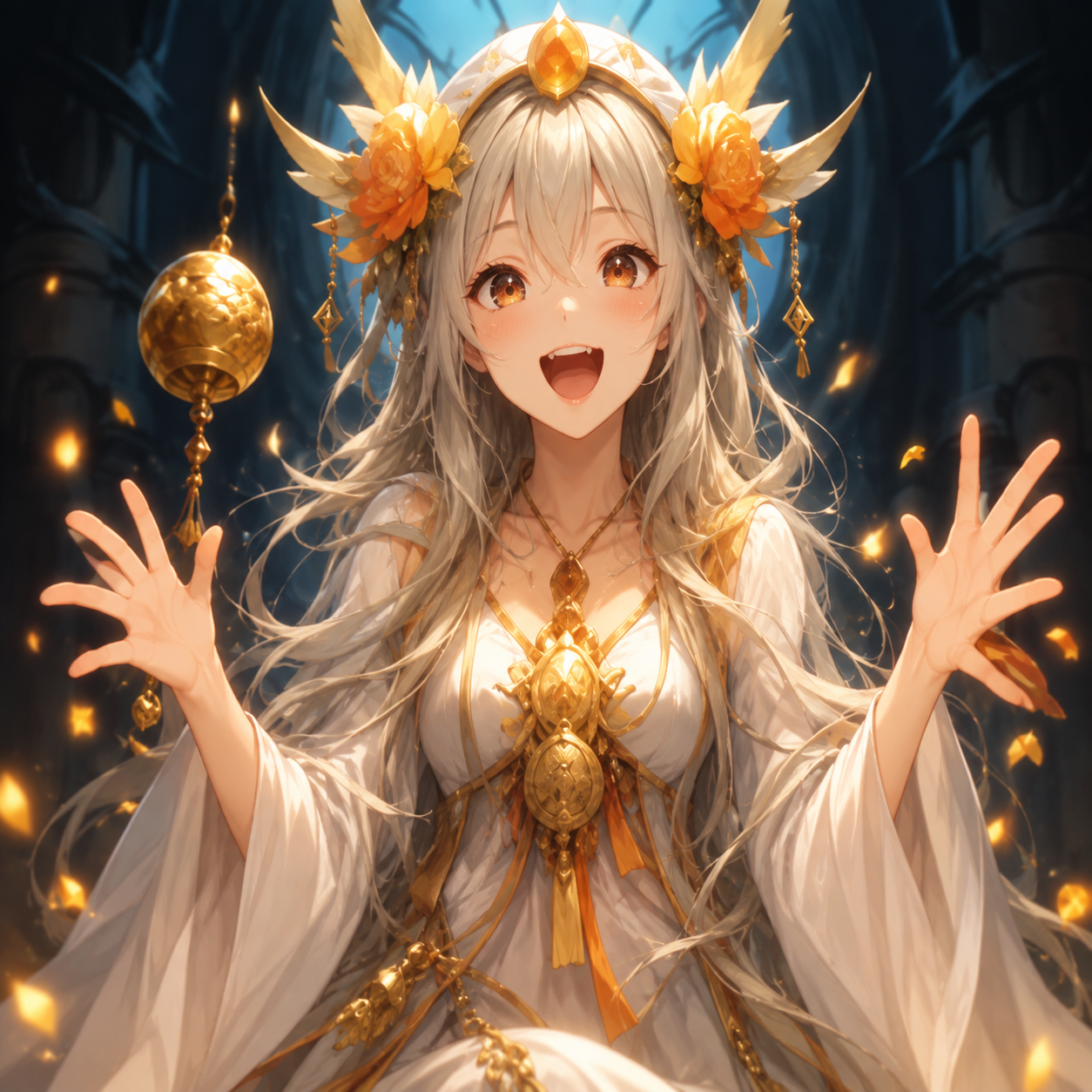} &
\qualimage{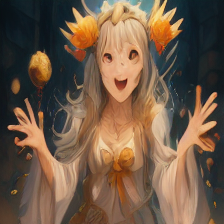} &
\qualimage{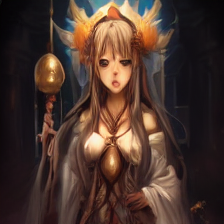} &
\qualimage{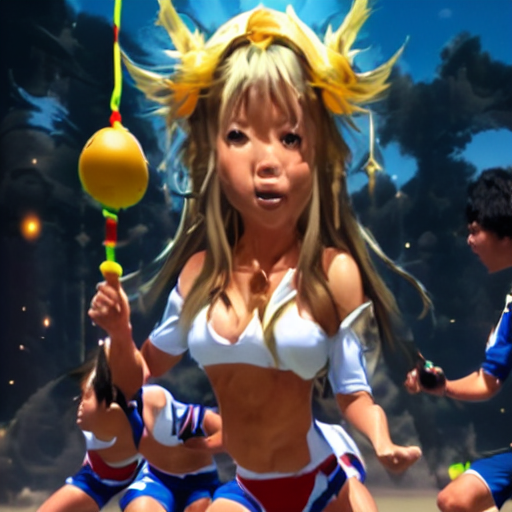} &
\qualimage{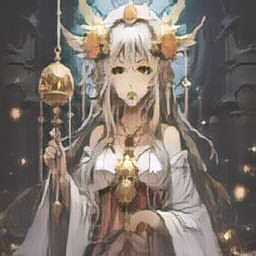} &
\qualimage{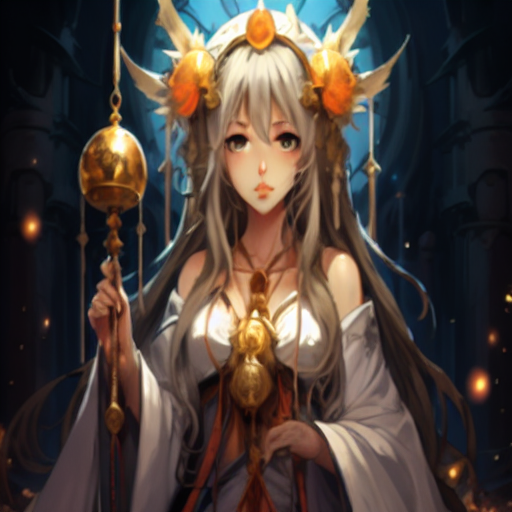} &
\qualimage{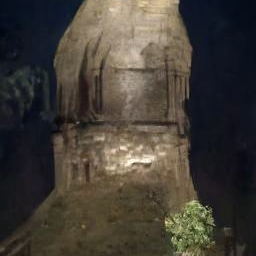} &
\qualimage{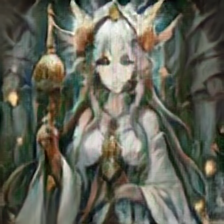}
\\
\noalign{\vskip1.5pt}

\qualtarget{qualTargetCool}{Fear} &
\qualimage{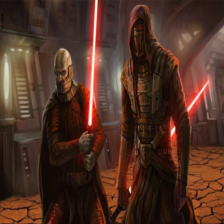} &
\qualimage{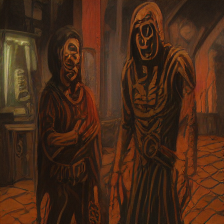} &
\qualimage{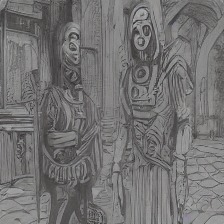} &
\qualimage{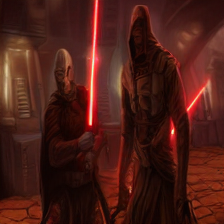} &
\qualimage{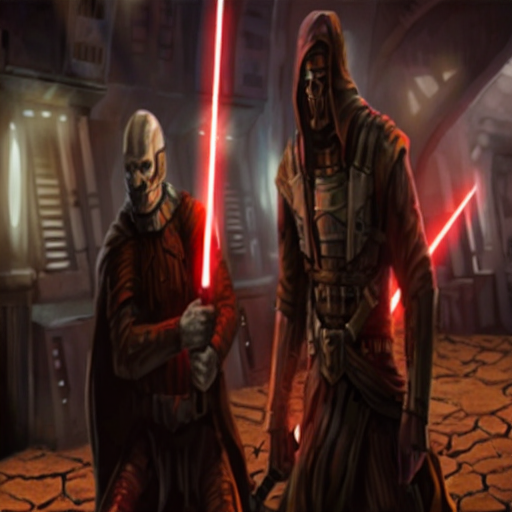} &
\qualimage{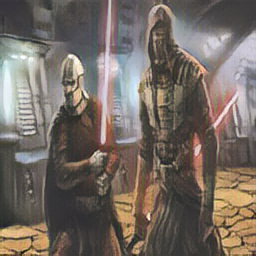} &
\qualimage{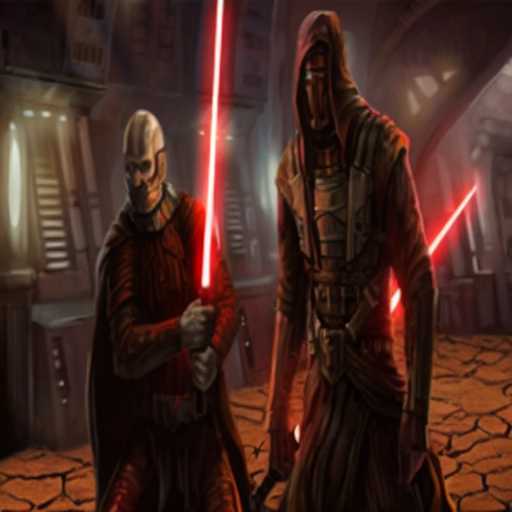} &
\qualimage{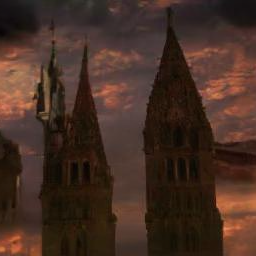} &
\qualimage{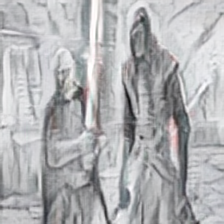}
\\
\noalign{\vskip1.5pt}

\qualtarget{qualTargetWarm}{Contentment} &
\qualimage{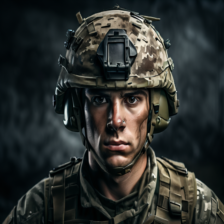} &
\qualimage{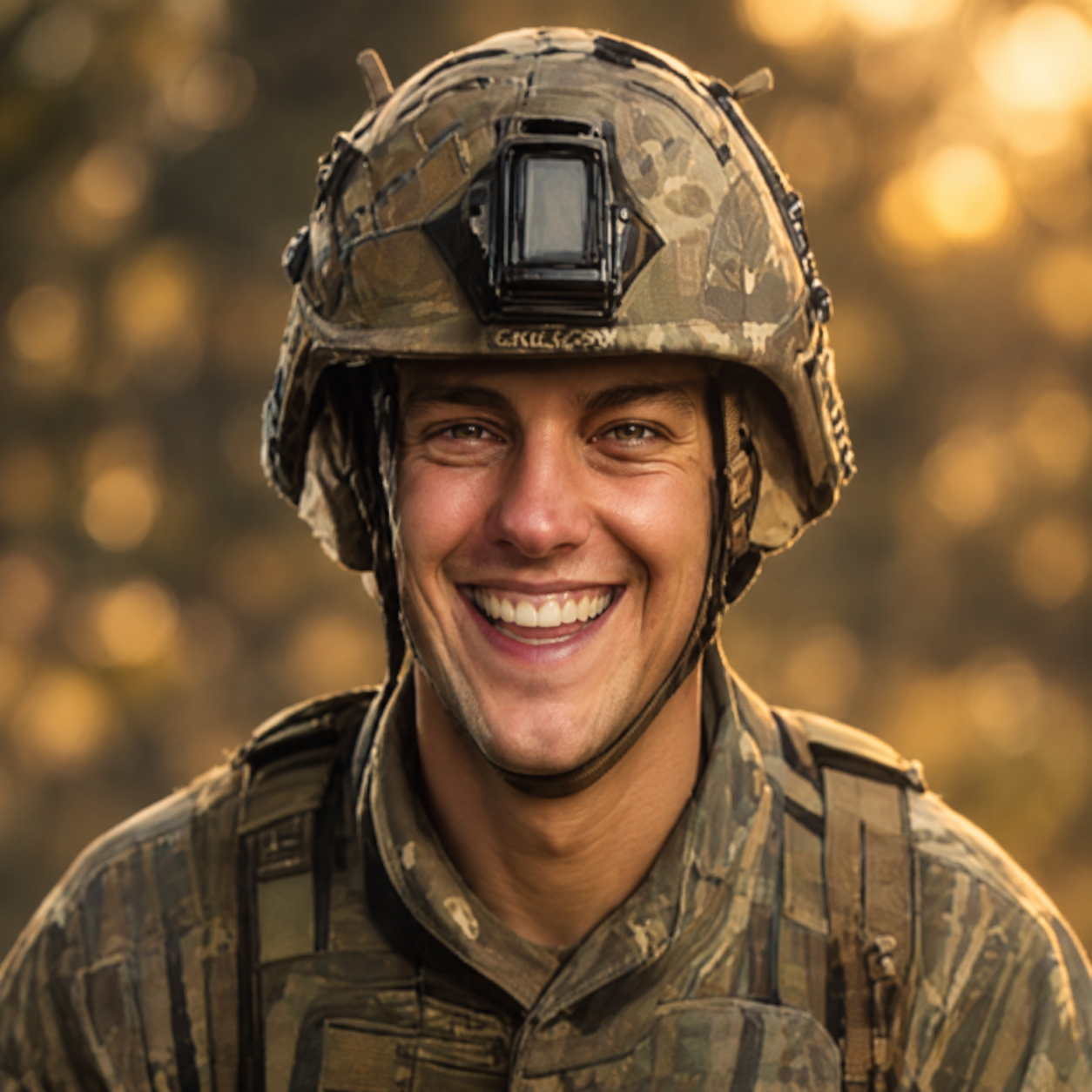} &
\qualimage{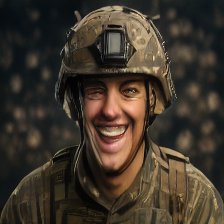} &
\qualimage{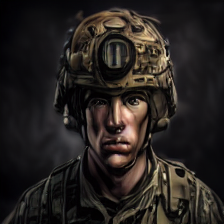} &
\qualimage{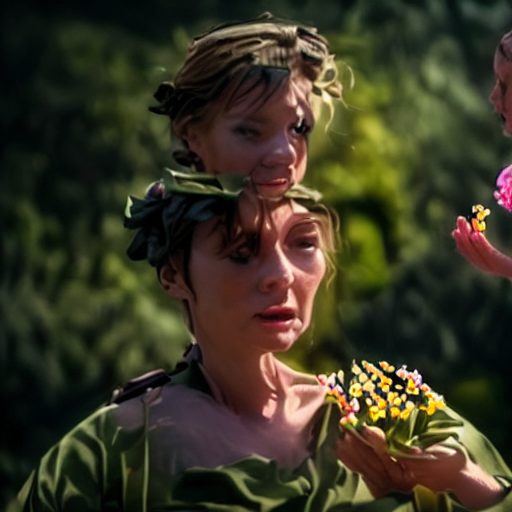} &
\qualimage{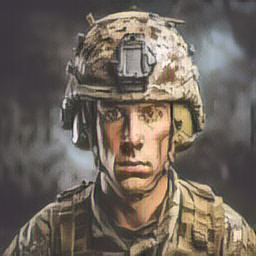} &
\qualimage{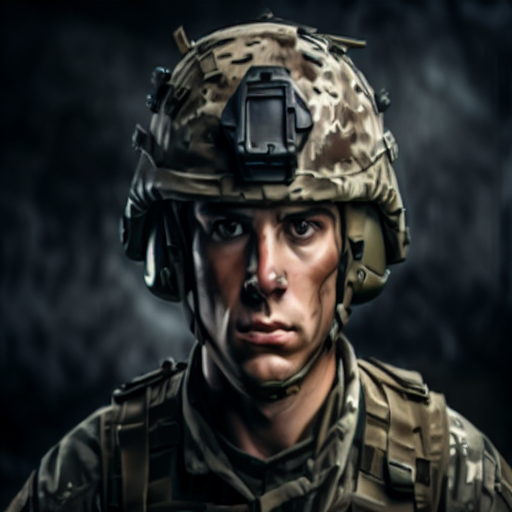} &
\qualimage{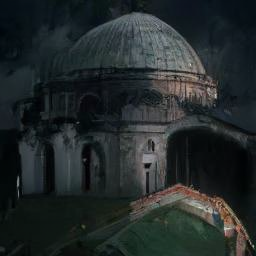} &
\qualimage{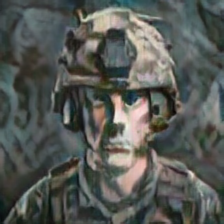}

\\
\noalign{\vskip1.5pt}

\qualtarget{qualTargetCool}{Disgust} &
\qualimage{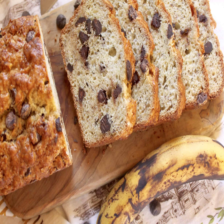} &
\qualimage{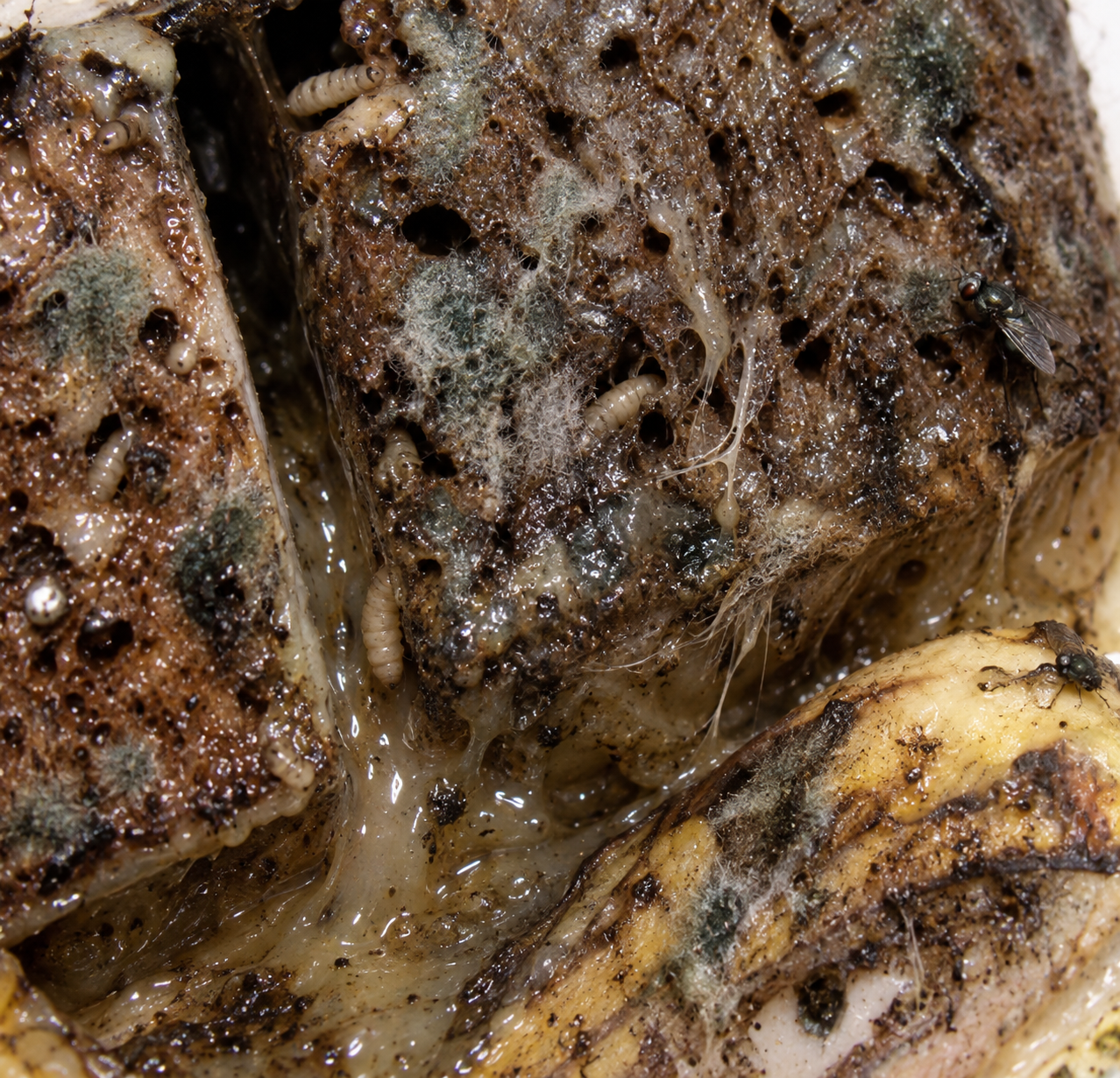} &
\qualimage{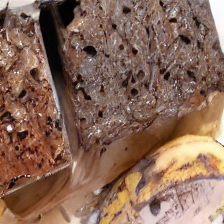} &
\qualimage{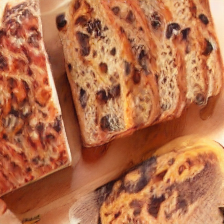} &
\qualimage{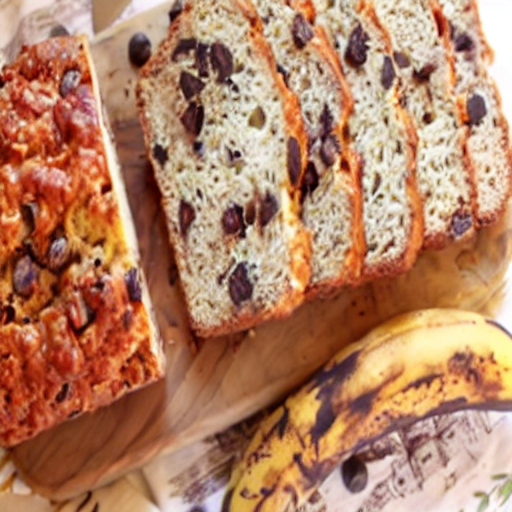} &
\qualimage{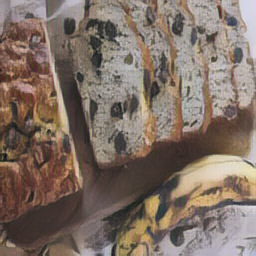} &
\qualimage{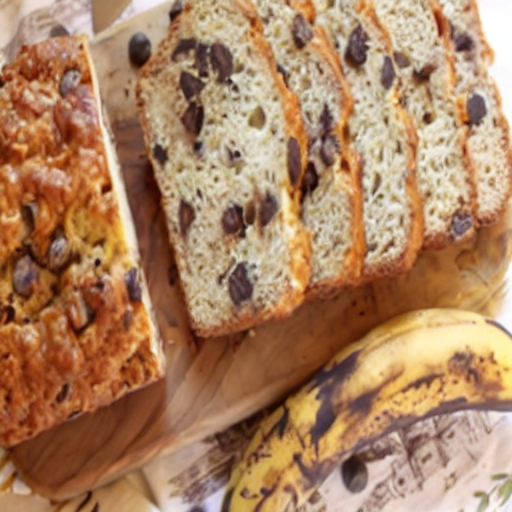} &
\qualimage{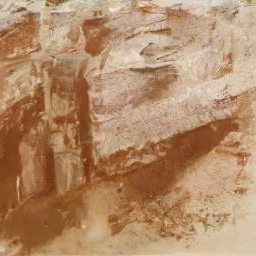} &
\qualimage{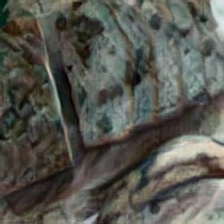}

\end{tabular}%
\endgroup
\caption{Qualitative comparison for eight target emotions. Each row shows the input image, the target emotion at left, and outputs from eight editing methods. Ours denotes ROI-ReFT with a ReFT write gain of 0.5, and AffectDelta is the base editor. IP2P denotes InstructPix2Pix.}
\label{fig:qualitative-comparison}
\vspace{-2mm}
\end{figure*}

\paragraph{Qualitative Comparison.}
\label{sec:qualitative}
Figure~\ref{fig:qualitative-comparison} presents one example for each of the eight target emotions. Each row uses the same source image and target emotion across methods. In the anger and excitement examples, ROI-ReFT changes facial expressions while retaining the man holding banknotes and the illustrated woman with her ornate headdress, robe, and staff, respectively. These edits illustrate how an emotional direction can be realized through scene-specific cues while keeping the principal subjects and their context recognizable. 

\subsection{Ablation Study}
\label{sec:ablation}

\paragraph{Ablation Settings.}
We compare seven settings in Table~\ref{tab:ablation} to examine neural guidance, interaction, and output supervision. ReFT w/o ROI retains the representation intervention but removes ROI guidance. Visual-only uses only the source visual condition, while Affective-only uses only the predicted affective change. Dual, independent lets each branch read its own ROI condition; Dual, interactive lets both branches read both conditions and constitutes the full model. Full w/o $\mathcal{L}_a$ removes affective response alignment on the generated image (Eq.~\ref{eq:affective_loss}), while Full w/o $\mathcal{L}_v$ removes the visual response budget (Eq.~\ref{eq:visual_budget}). These settings examine the source content reference and affective editing direction individually, jointly, and through their output objectives.

\begin{table}[htbp]
\centering
\caption{Ablation results on the same selected subset of 400 editing requests. Full denotes dual ROI guidance with interactive reading. All percentages use a denominator of 400; JSR additionally requires DINO-I $\geq 0.80$ for the same output. Best and second-best results in each column are bold and underlined, respectively, with ranks determined before rounding.}
\label{tab:ablation}
\begingroup
\small
\setlength{\tabcolsep}{3pt}
\renewcommand{\arraystretch}{1.12}
\begin{tabular*}{0.95\linewidth}{@{\extracolsep{\fill}}lcccccc@{}}
\toprule
Variant & Emo-A $\uparrow$ & $\Delta$Target $\uparrow$ & DINO-I $\uparrow$ & CLIP-I $\uparrow$ & LPIPS $\downarrow$ & JSR(0.80) $\uparrow$ \\
\midrule
ReFT w/o ROI & 78.500\% & $+0.231$ & 0.803 & 0.781 & 0.319 & 67.500\% \\
Visual-only & 83.000\% & $+0.246$ & \textbf{0.888} & \textbf{0.845} & \textbf{0.215} & 76.000\% \\
Affective-only & 89.000\% & $+0.289$ & 0.788 & 0.760 & 0.335 & 74.000\% \\
Dual, independent & 89.500\% & $+0.292$ & 0.837 & 0.801 & 0.277 & \underline{78.500\%} \\
\textbf{Dual, interactive} & \textbf{91.750\%} & $\underline{+0.301}$ & \underline{0.862} & \underline{0.819} & \underline{0.252} & \textbf{83.250\%} \\
\midrule
Full w/o $\mathcal{L}_a$ & 87.000\% & $+0.267$ & 0.858 & 0.817 & 0.257 & 77.250\% \\
Full w/o $\mathcal{L}_v$ & \underline{91.000\%} & $\mathbf{+0.308}$ & 0.805 & 0.773 & 0.319 & 75.500\% \\
\bottomrule
\end{tabular*}
\endgroup
\end{table}

\paragraph{Results and Analysis.}
The results support the complementary roles of the two neural conditions. Compared with ReFT w/o ROI, the full model raises Emo-A from 78.500\% to 91.750\% and JSR from 67.500\% to 83.250\%. Visual-only achieves the strongest preservation scores, whereas Affective-only achieves higher Emo-A (89.000\% versus 83.000\%) but lower JSR (74.000\% versus 76.000\%). This contrast supports using visual responses as a content reference and affective changes as an editing direction. Interactive reading further increases JSR over independent reading from 78.500\% to 83.250\%, while improving Emo-A and all three preservation metrics, supporting coordination between these roles. Output supervision helps realize this coordination: removing $\mathcal{L}_a$ lowers Emo-A to 87.000\% and JSR to 77.250\%, with comparatively small preservation changes. Removing $\mathcal{L}_v$ retains 91.000\% Emo-A and raises $\Delta$Target to $+0.308$, but weakens all three preservation scores and lowers JSR to 75.500\%. Thus, stronger target-emotion probability gains can accompany weaker content preservation. Together, the ablations support our motivation to let the source scene shape emotional editing and to supervise both the affective change and the visual content of the resulting image.

\subsection{Effect of Neural Guidance on Hidden Representations}
\label{sec:neural_correspondence}

To revisit the visual--affective imbalance in Figure~\ref{fig:motivation}b, we measure hidden-state correspondence with TRIBE-predicted target-image responses (Table~\ref{tab:neural_correspondence}). Random and shuffled $q_a$ test whether the affective direction is meaningful and paired with the source scene. Random affective ROI tests region selection by retraining $Q$ and ReFT with alternative regions, while Full w/o $\mathcal{L}_a$ tests affective output supervision. Appendix~\ref{sec:neural_control_details} details these controls.

\begin{table}[!htbp]
\centering
\caption{Hidden-state correspondence with TRIBE-predicted target-image responses, reported as Fisher $z$ scores. AffectDelta is the baseline in Figure~\ref{fig:motivation}b. The last column is the affective score minus 0.258. Best and second-best results are bold and underlined, respectively.}
\label{tab:neural_correspondence}
\begingroup
\small
\setlength{\tabcolsep}{4pt}
\renewcommand{\arraystretch}{1.12}
\begin{tabular*}{0.85\linewidth}{@{\extracolsep{\fill}}lcccc@{}}
\toprule
Configuration & \shortstack{Early visual\\$z\uparrow$} & \shortstack{Higher visual\\$z\uparrow$} & \shortstack{Affective ROI\\$z\uparrow$} & $\Delta z_{\mathrm{aff}}\uparrow$ \\
\midrule
AffectDelta & 0.427 & 0.447 & 0.258 & --- \\
Full + Random $q_a$ & 0.441 & 0.464 & 0.246 & $-0.012$ \\
Full + Shuffled $q_a$ & 0.448 & 0.460 & 0.262 & $+0.004$ \\
Random affective ROI & 0.453 & 0.470 & 0.284 & $+0.026$ \\
Full w/o $\mathcal{L}_a$ & \underline{0.468} & \underline{0.482} & \underline{0.352} & $\underline{+0.094}$ \\
\midrule
\textbf{ROI-ReFT Full} & \textbf{0.552} & \textbf{0.562} & \textbf{0.478} & $\mathbf{+0.220}$ \\
\bottomrule
\end{tabular*}
\vspace{-2mm}
\endgroup
\end{table}

\paragraph{Analysis.}
ROI-ReFT raises affective correspondence from 0.258 to 0.478, alongside gains in early and higher visual correspondence from 0.427 to 0.552 and from 0.447 to 0.562. The larger affective gain narrows the visual--affective gap that motivates our method. Norm-matched random conditions yield 0.246, and conditions borrowed from other scenes with similar emotional transitions yield 0.262, supporting a meaningful direction paired with the current scene. Random affective ROIs yield 0.284, supporting the selected affect-related regions over arbitrary regions of matched size. Removing $\mathcal{L}_a$ reduces affective correspondence to 0.352, demonstrating the contribution of output supervision. These representation-level findings complement the editing results: joint guidance strengthens affective correspondence while retaining strong visual correspondence.

\section{Conclusion}
\label{sec:conclusion}

We introduced ROI-ReFT to coordinate emotional change with source-content preservation. Our analysis found that the studied editors' hidden states correspond more strongly to TRIBE-predicted visual responses than to affect-related responses. Motivated by this finding, ROI-ReFT combines a source visual reference with a predicted affective response change. Two interacting branches guide the frozen editor throughout generation and learn from the generated image's predicted neural responses. On the post hoc selected subset of 400 editing requests, ROI-ReFT improves emotional accuracy and all three preservation metrics over AffectDelta, with 83.250\% joint success at DINO-I $\geq 0.80$. These results support using complementary neural guidance to shape how an image changes, connecting the requested emotion to the visual content that should remain recognizable.

\subsection*{AI Use Statement}
We used generative AI tools for literature review, manuscript polishing, and assistance with research code.     

\subsection*{Ethics Statement}
Our method uses existing image data and cortical responses predicted by a pretrained encoding model. These responses are computational estimates rather than direct measurements of an individual viewer's emotion. Emotional interpretations vary across people and contexts, and biases in the data and pretrained models may influence editing and evaluation. Applications should respect image rights and privacy, disclose emotional modifications, and avoid deceptive or coercive uses.

\subsection*{Reproducibility Statement}
Sections~\ref{sec:method} and~\ref{sec:experiments} describe the method and experimental comparisons. Appendices~\ref{sec:neural_encoding_details} and~\ref{sec:intervention_details} specify the neural encodings and intervention parameterization; Appendices~\ref{sec:loss_details} and~\ref{sec:training_inference} document the objectives, optimization, and inference settings. Metric definitions and neural guidance controls are detailed in Appendices~\ref{sec:evaluation_metrics} and~\ref{sec:neural_control_details}. The reported comparisons use the same post hoc selected set of 400 editing requests and shared frozen evaluators.

\FloatBarrier
\bibliography{references}
\bibliographystyle{iclr2027_conference}
\clearpage
\appendix
\section{Neural Guidance Encoding}
\label{sec:neural_encoding_details}

The frozen TRIBE v2 model predicts a population cortical response $B(I)\in\mathbb{R}^{20484}$. For a static image, we average the four retained response samples under a fixed presentation protocol. The visual state and affective change encoders summarize different quantities and use different normalization conventions.

For each of two visual ROI groups $\mathcal{V}_j$, the frozen state encoder standardizes, projects, and whitens the source response:
\begin{equation}
E_v^{\mathrm{state}}(B(I_s))=
\operatorname{concat}_{j=1}^{2}
\left[W_j^vP_j^v\left((B(I_s)_{\mathcal{V}_j}-\mu_j^v)\oslash\sigma_j^v\right)\right]
\in\mathbb{R}^{32}.
\label{eq:visual_encoding_details}
\end{equation}
Here $P_j^v$ maps each group to 16 coordinates, $W_j^v$ contains its coordinate normalization factors, and $\oslash$ denotes elementwise division. The source-dependent state is shared across the generation trajectory. The corresponding difference encoder $E_{v,j}^{\Delta}$ applies the same scaling, projection, and whitening to a response difference without subtracting $\mu_j^v$.

For three affect-related ROI groups $\mathcal{A}_k$, we encode a signed cortical response difference $\Delta b$ as
\begin{equation}
E_a(\Delta b)=\operatorname{concat}_{k=1}^{3}
\left[W_k^aP_k^a\left(\Delta b_{\mathcal{A}_k}\oslash\sigma_k^a\right)\right]
\in\mathbb{R}^{24}.
\label{eq:affective_code}
\end{equation}
Each group contributes eight coordinates. The group projections are fitted by uncentered singular value decomposition of training response differences, with vertex scaling and a single RMS scale per ROI group, represented by $W_k^a$. Encoding differences without mean subtraction preserves zero change and its sign, distinguishing this code from the centered visual state.

The predictor input $F_s\in\mathbb{R}^{256}$ is obtained by pooling the source VAE's unscaled posterior-mode latent to $4\times8\times8$. The predictor $Q$ applies layer normalization to $F_s$ and uses an MLP with two 256-dimensional hidden layers to map the concatenated source and emotion inputs to 24 coordinates. It is pretrained with group-averaged Smooth L1 supervision on paired affective change codes under its fixed response-prediction protocol and then frozen. We retain this pretrained encoding convention during intervention training. Final-image supervision recomputes source, target, and generated responses with a common continuous-image TRIBE protocol.

\section{Joint Intervention Parameterization}
\label{sec:intervention_details}

At layer $\ell$, timestep $t$, and spatial position $u$, let $h_{\ell,t,u}\in\mathbb{R}^{C_\ell}$ be the hidden vector and $\rho_{\ell,t,u}$ its channel RMS. A query combines the normalized hidden vector with source, time, and layer context:
\begin{equation}
\begin{aligned}
c_{\ell,t}&=S_\ell(F_s)+T_\ell(\tau_t)+e_\ell,\\
x_{\ell,t,u}&=H_\ell(h_{\ell,t,u}/\rho_{\ell,t,u})+c_{\ell,t}.
\end{aligned}
\label{eq:intervention_query}
\end{equation}
The time descriptor $\tau_t$ contains the normalized scheduler timestep, normalized log signal-to-noise ratio, and three smooth phase weights. A shared layer adapter can therefore change its action as denoising progresses.

Group-specific linear projections turn the three affective code groups into tokens $A_\ell$ and the two visual state groups into tokens $V_\ell$. Each token receives a learned group embedding. Separate scaled dot-product attention reads the two token sets, and a direct emotion projection provides the requested distribution change:
\begin{equation}
\begin{aligned}
a_{\ell,t,u}&=\operatorname{Attn}(x_{\ell,t,u},A_\ell,A_\ell),\\
v_{\ell,t,u}&=\operatorname{Attn}(x_{\ell,t,u},V_\ell,V_\ell),\\
e_\ell^p&=M_\ell([p_s,p_t,p_t-p_s]),\\
j_{\ell,t,u}&=[x_{\ell,t,u},a_{\ell,t,u},v_{\ell,t,u},c_{\ell,t},e_\ell^p].
\end{aligned}
\label{eq:joint_context}
\end{equation}
The trainable MLP $M_\ell$ complements the frozen affective predictor. Spatial selectivity comes from the queries formed from the editor's hidden states.

For each branch $b\in\{a,v\}$, an action network $f_\ell^b$ produces rank-$r$ coefficients, and an independent sigmoid gate $g_\ell^b$ controls their strength. The updates in Eq.~\eqref{eq:dual_write} are
\begin{equation}
\delta h_{\ell,t,u}^{b}
+=\operatorname{sg}(\rho_{\ell,t,u})\,
g_\ell^b(j_{\ell,t,u})\,
(\overline R_\ell^b)^\top f_\ell^b(j_{\ell,t,u}).
\label{eq:branch_write_details}
\end{equation}
Here $\overline R_\ell^b\in\mathbb{R}^{r\times C_\ell}$ is a trainable basis normalized rowwise. Each branch has its own basis, action network, and gate, while both use the joint context in Eq.~\eqref{eq:joint_context}. Their gates can activate simultaneously. Zero-initialized action output layers make the intervention initially preserve the base editor.

We apply the writes at the U-Net's \texttt{mid\_block} and \texttt{up\_blocks.2}, with 1280 and 640 channels, respectively. Each layer has a rank-16 adapter with token width 64. Both layers are active at every step of the 20-step DDIM trajectory. Conditions remain fixed for a given edit, while hidden states, queries, actions, and gates evolve. The original source latent remains concatenated with the noisy latent to retain the editor's spatial input.

\section{Output Distances and Auxiliary Objectives}
\label{sec:loss_details}

The visual distance in Eq.~\eqref{eq:visual_budget} averages the RMS differences of the two visual ROI codes:
\begin{equation}
d_v(I,I_s)=\frac{1}{2}\sum_{j=1}^{2}
\sqrt{\frac{\|E_{v,j}^{\Delta}(B(I)-B(I_s))\|_2^2}{16}+\epsilon}.
\label{eq:visual_distance}
\end{equation}
The paired target supplies the affective loss in Eq.~\eqref{eq:affective_loss}, while the matched base output $I_g^0$ supplies the reference deviation for the visual budget.

A frozen emotion predictor $P$ is supervised by $\mathcal{L}_e=\operatorname{KL}(p_t\|P(I_g))$. For content, define $d_C(I,J)=1-\cos(C(I),C(J))$, where $C$ is the frozen CLIP image encoder~\citep{radford2021learning}. We combine a budget on deviation from the matched base edit with a budget on source structure:
\begin{equation}
\mathcal{L}_c=
[d_C(I_g,I_g^0)-\kappa_C]_+^2
+\lambda_S[d_S(I_g,I_s)-\kappa_S]_+^2.
\label{eq:content_loss}
\end{equation}
The structural distance $d_S$ averages squared differences between normalized Sobel edge maps at resolutions $32$, $16$, and $8$. At each resolution, luminance is centered and RMS-normalized before extracting and normalizing the edges. This constrains coarse layout while accommodating brightness and contrast changes. We use $\kappa_C=0.05$, $\kappa_S=0.2$, and $\lambda_S=0.25$; each term penalizes only deviations exceeding its budget.

The noise-prediction objective is
\begin{equation}
\mathcal{L}_{\epsilon}=\mathbb{E}
\|\epsilon-\epsilon_{\theta,\phi}(z_t,I_s,p_s,p_t,t)\|_2^2,
\label{eq:noise_loss_details}
\end{equation}
where $z_t$ is a noisy latent of the paired target image $I_t$, timesteps cover the full training noise schedule, $\theta$ denotes the frozen editor, and $\phi$ the trainable intervention. The normalized update penalty $\mathcal{L}_u$ measures squared total write magnitude relative to squared hidden magnitude, averaged over the two intervention layers. On output-supervision updates, it averages this sampled-step penalty with the corresponding full-trajectory penalty.

\section{Training and Inference}
\label{sec:training_inference}

We optimize the intervention's token projections, context maps, emotion projections, action networks, gates, and per-layer bases. The base U-Net, VAE, emotion encoder, affective predictor $Q$, TRIBE, and ROI encoders remain frozen. Every update combines a randomly sampled noise-prediction objective with full-trajectory output supervision. Activation checkpointing retains gradient paths through all 20 denoising steps, the image decoder, and the frozen neural teacher.

Training uses the observed source-to-target emotion transitions to balance pair exposure, while retaining each pair's original images and soft emotion distributions. The sampler prioritizes underrepresented transitions and, when available within its balance tolerance, pairs consecutive selections from the same source with different target categories. We use AdamW for 2000 updates at a learning rate of $10^{-4}$, with four workers and one pair per worker, zero weight decay, and gradient clipping at 1. The loss weights are $\lambda_a=\lambda_v=1$, $\lambda_e=\lambda_c=0.1$, and $\lambda_u=0.001$; the visual ROI allowance is $\kappa_v=0.1$. The random seed is 20260918.

At inference, the source image supplies its latent features and visual response state, while $Q$ predicts the affective change from the source and requested emotion distributions. These conditions are computed once and reused throughout sampling. We use a uniform intervention gain $\gamma_a=\gamma_v=0.5$ for both layers and all steps in the current inference configuration. The gain scales the learned hidden-state writes in Eq.~\eqref{eq:dual_write}; training uses unit gain. A target image is required for paired training supervision but is unnecessary for editing a new source image.

\section{Evaluation Metrics}
\label{sec:evaluation_metrics}
We evaluate every method on the same $N=400$ source--target emotion pairs. For pair $i$, $I_s^i$ is the source image, $I_g^i$ is the edited output, and $y_t^i\in\{1,\ldots,8\}$ is the requested emotion category. The fixed emotion predictor from EmoEditor~\citep{lin2025make} returns the eight-class probability vector $P_{\mathrm{eval}}(I)$. Visual preservation is measured between $I_s^i$ and $I_g^i$. Each image pair contributes equally to the reported mean.

\subsection{Emotion Editing Metrics}
\paragraph{Emotion accuracy (Emo-A).}
An edit is emotionally successful when the evaluator's most probable category equals the requested target. Define
\begin{equation}
s_i=\mathbf{1}\!\left[\arg\max_{c}P_{\mathrm{eval}}(I_g^i)_c=y_t^i\right],
\qquad
\mathrm{Emo\text{-}A}=\frac{1}{N}\sum_{i=1}^{N}s_i.
\label{eq:metric_emo_a}
\end{equation}
The indicator is one for a correct target prediction and zero otherwise. We report Emo-A as a percentage, so each successful output contributes $100/N=0.250$ percentage points. This measures target-category correctness without imposing a visual preservation threshold.


\subsection{Visual Preservation Metrics}
\paragraph{DINO image similarity (DINO-I).}
Let $f_D(I)$ denote the image feature produced by the fixed DINO-based evaluation encoder, including its input preprocessing. We first normalize each feature to unit Euclidean norm, $\bar f_D(I)=f_D(I)/\|f_D(I)\|_2$, and compute source--output cosine similarity:
\begin{equation}
v_i=\bar f_D(I_s^i)^\top\bar f_D(I_g^i),
\qquad
\mathrm{DINO\text{-}I}=\frac{1}{N}\sum_{i=1}^{N}v_i.
\label{eq:metric_dino}
\end{equation}
Higher values indicate closer visual representations of the source and edited images. The joint-success calculation below uses each individual $v_i$ before averaging.

\paragraph{CLIP image similarity (CLIP-I).}
For the fixed CLIP image encoder~\citep{radford2021learning}, let $f_C(I)$ include its input preprocessing and image projection, and define $\bar f_C(I)=f_C(I)/\|f_C(I)\|_2$. CLIP-I is
\begin{equation}
c_i=\bar f_C(I_s^i)^\top\bar f_C(I_g^i),
\qquad
\mathrm{CLIP\text{-}I}=\frac{1}{N}\sum_{i=1}^{N}c_i.
\label{eq:metric_clip}
\end{equation}
This compares two images in CLIP's representation space, with higher scores indicating greater similarity. The source image is the reference for this evaluation, whereas the training content budget in Eq.~\eqref{eq:content_loss} uses the matched base edit for its CLIP term.

\paragraph{Learned Perceptual Image Patch Similarity (LPIPS).}
LPIPS~\citep{zhang2018perceptual} compares normalized intermediate feature maps using learned channel weights. Let $\hat\phi_\ell(I)_{hw}$ denote the channel-normalized feature at spatial position $(h,w)$ of layer $\ell$, and let $w_\ell$ be its learned channel weights. The per-pair distance and test-set mean are
\begin{equation}
\begin{aligned}
d_i^{\mathrm{LPIPS}}&=\sum_{\ell}\frac{1}{H_\ell W_\ell}\sum_{h=1}^{H_\ell}\sum_{w=1}^{W_\ell}
\left\|w_\ell\odot\left(\hat\phi_\ell(I_s^i)_{hw}-\hat\phi_\ell(I_g^i)_{hw}\right)\right\|_2^2,\\
\mathrm{LPIPS}&=\frac{1}{N}\sum_{i=1}^{N}d_i^{\mathrm{LPIPS}}.
\end{aligned}
\label{eq:metric_lpips}
\end{equation}
Here $H_\ell$ and $W_\ell$ are the feature-map dimensions and $\odot$ denotes elementwise multiplication. Smaller distances indicate greater perceptual similarity. Unlike DINO-I and CLIP-I, LPIPS aggregates spatial feature differences across multiple layers.

\subsection{Joint Success Rate (JSR)}
JSR measures how often the same edited image achieves the emotional goal and meets a specified visual preservation requirement. Using the emotion-success indicator $s_i$ from Eq.~\eqref{eq:metric_emo_a} and the individual DINO similarity $v_i$ from Eq.~\eqref{eq:metric_dino}, we define
\begin{equation}
\mathrm{JSR}(\tau)=\frac{1}{N}\sum_{i=1}^{N}s_i\,\mathbf{1}[v_i\ge\tau],
\qquad N=400.
\label{eq:metric_jsr}
\end{equation}
The denominator is always the entire test set. An output contributes only if both conditions hold for that output; success on different images cannot be combined. Thus, JSR evaluates the intersection of emotional correctness and visual preservation rather than a product of their separate average scores.

All methods use the same 400 pairs and the same thresholds. For every $\tau$, we compare each unrounded per-image DINO similarity with $\tau$, count the jointly successful outputs, and divide by 400. We report $100\,\mathrm{JSR}(\tau)$ on the percentage axis of Figure~\ref{fig:jsr-comparison}, sweeping the displayed threshold range $[0,0.90]$. For example, ROI-ReFT has 333 jointly successful outputs at $\tau=0.80$, giving $333/400=0.8325$, reported as 83.250\%.

For fixed outputs, JSR is non-increasing as $\tau$ increases and is bounded above by Emo-A. When a threshold admits every image pair, JSR equals Emo-A. The curve therefore shows how much emotional success remains as the required visual similarity becomes stricter. Table values are rounded to three decimal places for presentation; thresholding is performed before rounding.


\section{Neural Guidance Control Settings}
\label{sec:neural_control_details}

The controls in Table~\ref{tab:neural_correspondence} distinguish the information carried by the affective condition, its pairing with the source scene, and the choice of guidance regions.

\paragraph{Random $q_a$.}
For each request, we replace $\widehat q_a=Q(F_s,p_s,p_t,p_t-p_s)$ with a random 24-dimensional vector generated in the training-set standardized coordinates and matched to the norm of the predicted condition it replaces. We keep the source image, visual condition $v_s$, emotion inputs, diffusion noise, and model weights unchanged. Results are summarized over multiple predefined random seeds per request. Matching the norm separates direction information from changes in input magnitude.

\paragraph{Shuffled $q_a$.}
For request $i$, we select another request $j\ne i$ with the same target emotion, a similar source emotion distribution and transition magnitude, and a different source scene. We use request $j$'s predicted affective condition while retaining request $i$'s source image, $v_s$, $p_s$, $p_t$, diffusion noise, and model weights. This control preserves a genuine output of $Q$ and approximately preserves emotional-transition information while disrupting its pairing with the current scene.

\paragraph{Random affective ROI.}
We select regions outside the original affective ROIs with matching vertex counts, reconstruct the 24-dimensional affective code using the same encoding procedure, and retrain both $Q$ and ReFT. The visual ROIs remain unchanged. We repeat the experiment with multiple random region selections to examine whether the benefit depends on the chosen affect-related regions.

\paragraph{Removing affective output supervision.}
Full w/o $\mathcal{L}_a$ retains both guidance branches and removes the affective response alignment objective on the generated image, as in Section~\ref{sec:ablation}.

\section{Hidden-State Correspondence Analysis of Existing Editors}
\label{sec:baseline_hidden_correspondence}

This analysis provides the encoding protocol and controls for the EmoEdit and EmoEditor results in Figure~\ref{fig:motivation}b. We examine how well an editor's intermediate representation predicts the neural response associated with its paired target image.

\paragraph{Data and representation extraction.}
The two editors use the same 400 source--target pairs in the same order. Source and target emotion categories are each balanced across eight classes, with 50 samples per category. In 155 pairs, source and target share the same dominant emotion category; the remaining 245 pairs change category. Each editor receives the source image and target emotion. We record the output of \texttt{unet.mid\_block} at step 15 of a 30-step generation trajectory. This diagnostic extraction schedule is separate from the 20-step inference configuration of ROI-ReFT. TRIBE v2 predicts responses to the source and paired target images; the encoding analysis uses the paired target response as its prediction target.

\paragraph{ROI encoding and controls.}
We analyze five ROI groups: early visual (V1/V2/V3), V4/higher visual, affect-related, auditory, and sensorimotor regions. For each editor and ROI group, we fit a separate linear encoding model using a five-fold pipeline of feature standardization, principal component analysis, and cross-validated ridge regression. The ridge objective is
\begin{equation}
W^\star=\arg\min_W\|Y-XW\|_F^2+\alpha\|W\|_F^2,
\label{eq:baseline_roi_probe}
\end{equation}
where $X$ contains the transformed hidden features and each row of $Y$ contains the paired target image's predicted response within ROI $\mathcal{R}$, $Y_{i,:}=B(I_t^i)|_{\mathcal{R}}$. Thus, the probe predicts a target response state, whereas the affective guidance in our method specifies a response change. We evaluate predictions on held-out folds using Pearson correlation and report Fisher-transformed scores, $z=\operatorname{atanh}(r)$.

We compare the actual hidden features with two controls. \emph{Shuffled hidden} permutes hidden-to-response pairings within folds among requests with the same target emotion. This preserves target-category membership while disrupting the correspondence between individual requests and their paired target responses. \emph{Random features} uses Gaussian features with the same shape as the hidden features. Both controls repeat the complete encoding pipeline and five-fold evaluation. These controls operate on the features used for the probe, whereas the $q_a$ controls in Section~\ref{sec:neural_correspondence} perturb the neural guidance supplied to ROI-ReFT.

\begin{table}[!htbp]
\centering
\caption{Hidden-state correspondence with TRIBE-predicted paired-target responses for existing editors. Scores are Fisher-transformed Pearson correlations. Shuffled hidden preserves target-emotion groups while disrupting pairings; Random features replaces hidden features with same-shaped Gaussian features.}
\label{tab:baseline_roi_controls}
\begingroup
\small
\setlength{\tabcolsep}{4pt}
\renewcommand{\arraystretch}{1.1}
\begin{tabular*}{\linewidth}{@{\extracolsep{\fill}}lrrrrrr@{}}
\toprule
& \multicolumn{3}{c}{EmoEditor} & \multicolumn{3}{c}{EmoEdit} \\
\cmidrule(lr){2-4}\cmidrule(lr){5-7}
ROI group & Actual & Shuffled & Random & Actual & Shuffled & Random \\
\midrule
Early visual & 0.391 & 0.016 & $-0.015$ & 0.405 & 0.051 & $-0.039$ \\
V4/higher visual & 0.460 & 0.032 & $-0.039$ & 0.356 & 0.070 & $-0.017$ \\
Affective & 0.240 & 0.007 & $-0.035$ & 0.237 & 0.132 & 0.001 \\
Auditory control & 0.232 & 0.034 & 0.074 & 0.187 & 0.105 & 0.092 \\
Sensorimotor control & 0.206 & $-0.014$ & $-0.021$ & 0.191 & 0.106 & $-0.001$ \\
\bottomrule
\end{tabular*}
\endgroup
\end{table}

\paragraph{Results.}
Actual features exceed both controls in every ROI group (Table~\ref{tab:baseline_roi_controls}), indicating correspondence with the predicted responses beyond the tested pairing and random-feature controls. Both editors have their two highest scores in visual regions. EmoEditor's highest visual score is 0.460, compared with 0.240 in affect-related regions; EmoEdit's corresponding scores are 0.405 and 0.237. Higher scores than the same-target shuffled control indicate that the correspondence depends on request-level pairing beyond target-category membership. Positive correspondence also appears in auditory and sensorimotor regions. The stronger visual correspondence and comparatively weaker affective correspondence reveal the representational imbalance that motivates the joint guidance design studied in the main paper.


\end{document}